\documentclass[11pt]{article}

\usepackage{acl}

\usepackage{times}
\usepackage{latexsym}

\usepackage[T1]{fontenc}

\usepackage[utf8]{inputenc}

\usepackage{microtype}

\usepackage{inconsolata}

\usepackage{graphicx}

\usepackage{amsmath}
\usepackage{amsfonts}
\usepackage{nicefrac}
\usepackage{booktabs}
\usepackage{enumitem}
\usepackage{xspace}
\usepackage{colortbl}
\usepackage{multirow}
\usepackage{subcaption}
\usepackage{algorithm}
\usepackage{algpseudocode}
\usepackage{makecell}

\usepackage[table]{xcolor}
\usepackage{array}

\usepackage{xcolor}
\usepackage{tabularx}

\definecolor{humangray}{HTML}{999999}
\definecolor{baseblue}{HTML}{0072B2}
\definecolor{alignedvermillion}{HTML}{D55E00}
\definecolor{pastagreen}{HTML}{009E73}
\definecolor{fluencycolor}{HTML}{3E5C76}
\definecolor{relevancecolor}{HTML}{C97064}

\newcolumntype{A}{>{\columncolor{alignedvermillion!12}}r}
\newcolumntype{P}{>{\columncolor{pastagreen!12}}r}

\newcommand{\legendbox}[2]{%
  \begingroup
  \setlength{\fboxsep}{1.5pt}%
  \colorbox{#1}{\textcolor{white}{\scriptsize\textbf{#2}}}%
  \endgroup
}

\newcommand{\pasta}{PASTA\xspace}

\newcommand{\ClaudeSonnet}{Claude Sonnet 4.6\xspace}

\newcommand{\Pangram}{Pangram\xspace}
\newcommand{\GPTZero}{GPTZero\xspace}

\newcommand{\RAIDAR}{RAIDAR\xspace}

\title{Measuring, Localizing, and Reducing the AI-Like Writing Style of Post-Trained LLMs}

\makeatletter
\renewcommand{\@fnsymbol}[1]{\ifcase#1\or =\or \ensuremath{\dagger}\or \ensuremath{\ddagger}\else\@ctrerr\fi}
\makeatother
\newcommand*\samethanks[1][\value{footnote}]{\footnotemark[#1]}

\author{
  \textbf{Aniket Anand}\textsuperscript{1}\thanks{Equal contribution.}\enskip
  \textbf{Janvijay Singh}\textsuperscript{2}\samethanks[1]\enskip
  \textbf{Zhewei Sun}\textsuperscript{3}\enskip
  \textbf{Dilek Hakkani-T\"ur}\textsuperscript{2}\enskip
  \textbf{Nick Feamster}\textsuperscript{1}
  \\[0.5em]
  \textsuperscript{1}University of Chicago \quad
  \textsuperscript{2}University of Illinois Urbana-Champaign
  \\
  \textsuperscript{3}Toyota Technological Institute at Chicago
  \\
  {\small
  \texttt{\{aanand300, feamster\}@uchicago.edu},\;
  \texttt{zsun@ttic.edu},\;
  \texttt{\{jvsingh2, dilek\}@illinois.edu}}
}

\begin{document}
\maketitle
\begin{abstract}
Post-trained language models often exhibit a recognizable AI-like writing style, yet its connection to post-training and internal representations remains poorly understood.
In this work, we study whether post-training introduces or amplifies AI-like stylistic regularities and whether these regularities have a localized internal signature.
To this end, we compare human text, base-model generations, and aligned-model generations under matched human-source prefixes. 
Aligned generations show lower human-corpus affinity and higher AI-detection rates than base generations, suggesting that post-training shifts generated text away from human-corpus style and toward detector-visible AI-like text.
We then introduce \textbf{PASTA} (\textbf{P}ost-training \textbf{A}lignment \textbf{S}ignature \textbf{T}argeted \textbf{A}blation), a training-free method that estimates a post-training alignment signature from aligned-base residual contrasts and ablates the corresponding direction during decoding.
Across 11 aligned models and 6 AI-detectors, PASTA lowers the detection rate for most aligned models; this effect transfers well across detectors and is not reproduced by random directions.
Qualitative analysis suggests that PASTA generations remain relevant and coherent while exhibiting greater stylistic variation.
Together, these results show that AI-like stylistic effects of post-training can be measured, localized, and causally tested through activation ablation.\footnote{\textbf{Code+Data:} \href{https://github.com/aanand300/pasta}{github.com/aanand300/pasta}.}

\end{abstract}

\section{Introduction}

Post-trained language models produce text with a recognizable style: polished phrasing, hedging, affirming phrases, and highly regular structure~\citep{chakrabarty2025salvaged,russell2026storyscope}.
While these patterns are familiar at the output level, they raise a deeper question:
\textit{To what extent are AI-like stylistic regularities introduced by post-training, and how are they represented inside the model?}
Prior work has shown that post-training changes output diversity, expressed opinions, verbosity, and behaviors such as sycophancy~\citep{kirk2024rlhfdiversity,santurkar2023opinions,perez2023discovering,sharma2024sycophancy}.
However, these studies characterize post-training effects behaviorally, focusing on output-level changes.
It remains unclear whether the recognizable AI-like writing style itself is a post-training effect and whether a meaningful component of this effect can be localized in, and ablated from, aligned-model representations.

To address this gap, we contrast aligned models with their base counterparts under matched prompting conditions. This contrast isolates stylistic regularities introduced or amplified by post-training from those already present after pretraining. We then ask whether this post-training effect is only an output-level distributional shift, or whether it corresponds to an identifiable structure inside the model. If the effect is diffuse across representations, it may be difficult to localize or intervene on. If, instead, a meaningful component is captured by a single residual-stream direction, then post-training leaves a concrete representational footprint. We call this direction a \emph{post-training alignment signature}. Building on work that links behaviors such as refusal to linear directions in representation space~\citep{arditi2024refusal,zou2023representation}, we ask whether broader AI-like stylistic regularities induced by post-training can be localized and causally tested in the same way.

This leads to two empirical questions: 1) Does post-training induce an output-level shift from base to aligned generations? 2) If a shift exists, can it be localized to an identifiable direction in the residual stream? To answer the first question, we use matched prefixes to compare human text, base-model generations, and aligned-model generations. 
Aligned generations show lower affinity to human-corpus indexes under infini-gram~\citep{liu2024infini} and higher AI-detection rates than base generations, suggesting that post-training makes generated text less human-corpus-like and more detector-visible. We use detectors as probes of stylistic separability rather than training objectives.

To answer the second question, we introduce \textbf{PASTA} (\textbf{P}ost-training \textbf{A}lignment \textbf{S}ignature \textbf{T}argeted \textbf{A}blation), a training-free method that estimates an aligned-minus-base residual direction and removes it during aligned-model decoding. A detector-guided model layer sweep selects the direction used for ablation.
Across 11 aligned models and six detectors, PASTA lowers detector flagging for most models, with  Pangram~\citep{emi2024technical} reductions exceeding 90 percentage points for several models. The effect transfers to detectors not used for layer selection and is not reproduced by random directions, suggesting that the extracted direction captures a specific post-training signal. 
Qualitative analysis suggests that PASTA generations remain relevant and coherent while exhibiting greater stylistic variation.

Our contributions are twofold. 
First, we provide output-level evidence that post-training shifts aligned generations away from human-corpus style and toward detector-visible AI-like text. 
Second, we provide evidence that post-training leaves an identifiable representation-level signature associated with AI-like stylistic regularities in generated text. 
We introduce PASTA, a training-free intervention that extracts this signature from aligned-base activation contrasts and projects it out during aligned-model decoding.
Together, these results show that AI-like writing style induced by post-training can be measured, localized, and causally tested through interventions on model activations.

\begin{figure*}[t]
    \centering
    \includegraphics[width=\textwidth]{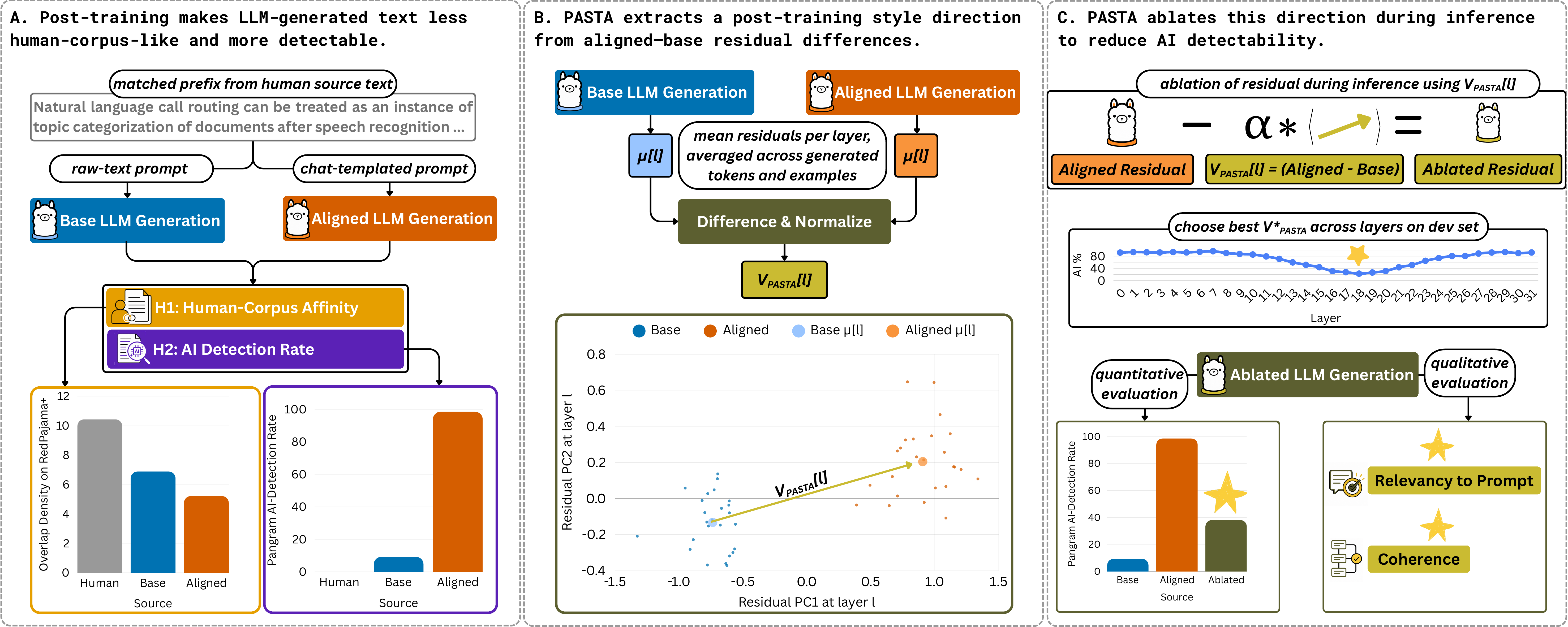}

\caption{
\textbf{PASTA overview.}
\textbf{(A)} Under matched human-sourced prefixes, aligned-model generations show lower infini-gram affinity to RedPajama+ and higher AI-detection rates than base-model generations, suggesting that post-training shifts generated text away from human-corpus style and toward detector-visible text.
\textbf{(B)} PASTA estimates this shift as a layer-wise residual-stream direction by averaging residual states over generated tokens and examples, then taking the normalized aligned-minus-base difference.
\textbf{(C)} A detector-guided layer sweep selects the most effective ablation direction, which PASTA projects out during inference. The resulting ablated generations have lower AI detectability while maintaining comparable prompt relevance and coherence in qualitative evaluations.
}
    
    \label{fig:pasta-overview}
\end{figure*}

\section{Related Work}
\label{sec:related_work}

\paragraph{Activation-Space Interventions and Post-Training Shifts.}
One line of work shows that behaviors can be localized and edited in
activation space. Representation Engineering~\citep{zou2023representation}, Activation
Engineering~\citep{turner2023actadd}, Inference-Time Intervention~\citep{li2023iti},
and Contrastive Activation Addition~\citep{rimsky2024caa} steer model behavior
at inference time by adding or removing components along activation
directions, without updating weights. These methods have been used to
steer behaviors such as truthfulness, sentiment, and compliance.
Most closely related, \citet{arditi2024refusal} identify a single residual-stream
direction mediating refusal behavior in aligned models.
A second line of work shows that post-training reshapes model outputs beyond
task performance. Instruction tuning and RLHF~\citep{ouyang2022training,
bai2022constitutional} affect expressed opinions and political
leaning~\citep{santurkar2023opinions}, sycophancy and reward-driven
behaviors~\citep{perez2023discovering,sharma2024sycophancy,
wei2023synthsycophancy}, verbosity and length bias~\citep{saito2023verbosity,
singhal2024length}, output diversity~\citep{kirk2024rlhfdiversity}, and
recognizable writing-style artifacts~\citep{chakrabarty2025salvaged,
russell2026storyscope}. 
These studies primarily characterize post-training effects behaviorally. 
Our work connects these two lines of work:
we ask whether an AI-like stylistic effect of post-training is not only
visible in outputs but also localized in model activations and causally testable
through ablation.

\paragraph{Base--Aligned Contrasts and AI-Text Detectors.}
Prior work compares base and aligned variants to isolate what post-training
adds. \citet{lin2024urial} show that prompting can recover much of aligned
models' surface behavior from base models; \citet{fei2025nudging} switch
between base and aligned decoding using base-model uncertainty;
\citet{gui2024bonbon} analyze best-of-$n$ alignment as a proxy for
post-training effects; and \citet{wang2025basealigned} mix base and aligned
generations to recover output diversity. We also use the base model as a
reference point, but focus on localizing the post-training shift rather than
mixing or switching between models.
We use AI-text detectors as measurement probes for this shift. Existing
detectors include likelihood-based methods~\citep{mitchell2023detectgpt,
bao2024fastdetect}, observer-model approaches such as
Binoculars~\citep{hans2024binoculars}, rewriting-based methods such as
RAIDAR~\citep{mao2024raidar}, and watermarking~\citep{kirchenbauer2023watermark}.
Since detector behavior varies across models, decoding settings, and
domains~\citep{tufts2024practical}, we treat detector outputs as signals of
stylistic separability, not ground-truth authorship labels. Prior work has also
shown that paraphrasing, rewriting, and stylometric obfuscation can reduce
detector confidence
\citep{krishna2023dipper,tripto2024shiptheseus,xing2024alison,lu2024sico,cheng2025advparaphrase,fang2025contrastpara}.
In contrast, we
use reduced detector visibility to study a post-training representation-level
signature, not to claim robust detector evasion.

\section{Post-Training Shifts Generations Away from Human-Corpus Text}
\label{sec:posttraining}

Post-training optimizes base models toward a much narrower text distribution
than web-scale pretraining: instruction and preference data are curated for
helpful, polished, structured, and rater-preferred responses. We ask whether
this narrower optimization target leaves a surface-level signature in
generation, beyond improving instruction following. Specifically, \textit{does
post-training move outputs away from naturally occurring human text and toward
a more recognizable assistant-like style?}

\paragraph{Setup.}
We test the post-training shift with matched triples:
\[
(t_{\mathrm{human}},\; t_{\mathrm{base}},\; t_{\mathrm{aligned}}),
\]
constructed for each source domain and base/aligned model pair. The domains are
scientific abstracts, creative fiction, opinion pieces, college essays, and
news articles. Each model pair consists of a pretrained base model and its
post-trained counterpart, such as Llama-3.1-8B and Llama-3.1-8B-Instruct.
For each triple, $t_{\mathrm{human}}$ is the original human continuation, while
$t_{\mathrm{base}}$ and $t_{\mathrm{aligned}}$ are generated from the same
human-written prefix under matched prompting. To avoid length as a confound, we length-control all continuations before
evaluation: base and aligned generations are retained only if their character
length is within $\pm 20\%$ of the corresponding human continuation. By fixing
the prefix, domain, model family, and length, the base--aligned contrast
isolates the effect of post-training. Appendix~\ref{app:prefix-extraction}
details the domain sources and prefix construction; Appendix~\ref{app:triplet-gen-settings}
details the models and generation settings; Appendix~\ref{app:prompts} details
prompt formats.

\paragraph{Approach.}
We probe the post-training surface shift with two complementary hypotheses.
\textbf{H1} tests whether aligned outputs have lower verbatim affinity to a
pretraining-style human reference corpus than base outputs. \textbf{H2} tests
whether aligned outputs receive higher AI-detector scores than base outputs.
These probes are not complete measures of human-likeness; they capture complementary properties. Corpus affinity measures proximity to naturally occurring human text, while detector scores capture detector-visible AI-generated regularities.

\paragraph{H1: Post-training reduces verbatim source affinity.}
Our first hypothesis tests whether post-training moves model generations away from naturally
occurring human text. We measure this using verbatim affinity
to a large reference corpus. For a reference corpus $C$ and a text
$t=(w_1,\ldots,w_N)$, let $\ell_i$ be the length, in words, of the longest
contiguous span beginning at position $i$ that appears at least once in $C$.
We define the \emph{overlap density} of $t$ as
\begin{equation}
\rho_C(t) = \frac{1}{N} \sum_{i=1}^{N} \ell_i,
\label{eq:infini-gram-density}
\end{equation}
where higher values indicate longer corpus-matched spans across the text. We
also report Coverage@10, the fraction of words covered by corpus-matched spans
of at least 10 words, as a long-span robustness check.

Let $\rho_{\mathrm{H}}$, $\rho_{\mathrm{B}}$, and $\rho_{\mathrm{A}}$ denote
the mean overlap density for human, base, and aligned continuations,
respectively. H1 predicts
\begin{equation}
\rho_{\mathrm{H}} > \rho_{\mathrm{B}} > \rho_{\mathrm{A}}.
\label{eq:verbatim-affinity}
\end{equation}
The first inequality reflects that human continuations are drawn from the
source distribution. The second is the key post-training test: if post-training
pushes models toward a narrower assistant-style distribution, aligned
generations should have lower affinity to pretraining-style human text than
base generations from the same prefixes.

\begin{figure}[t]
\centering
\includegraphics[width=\linewidth]{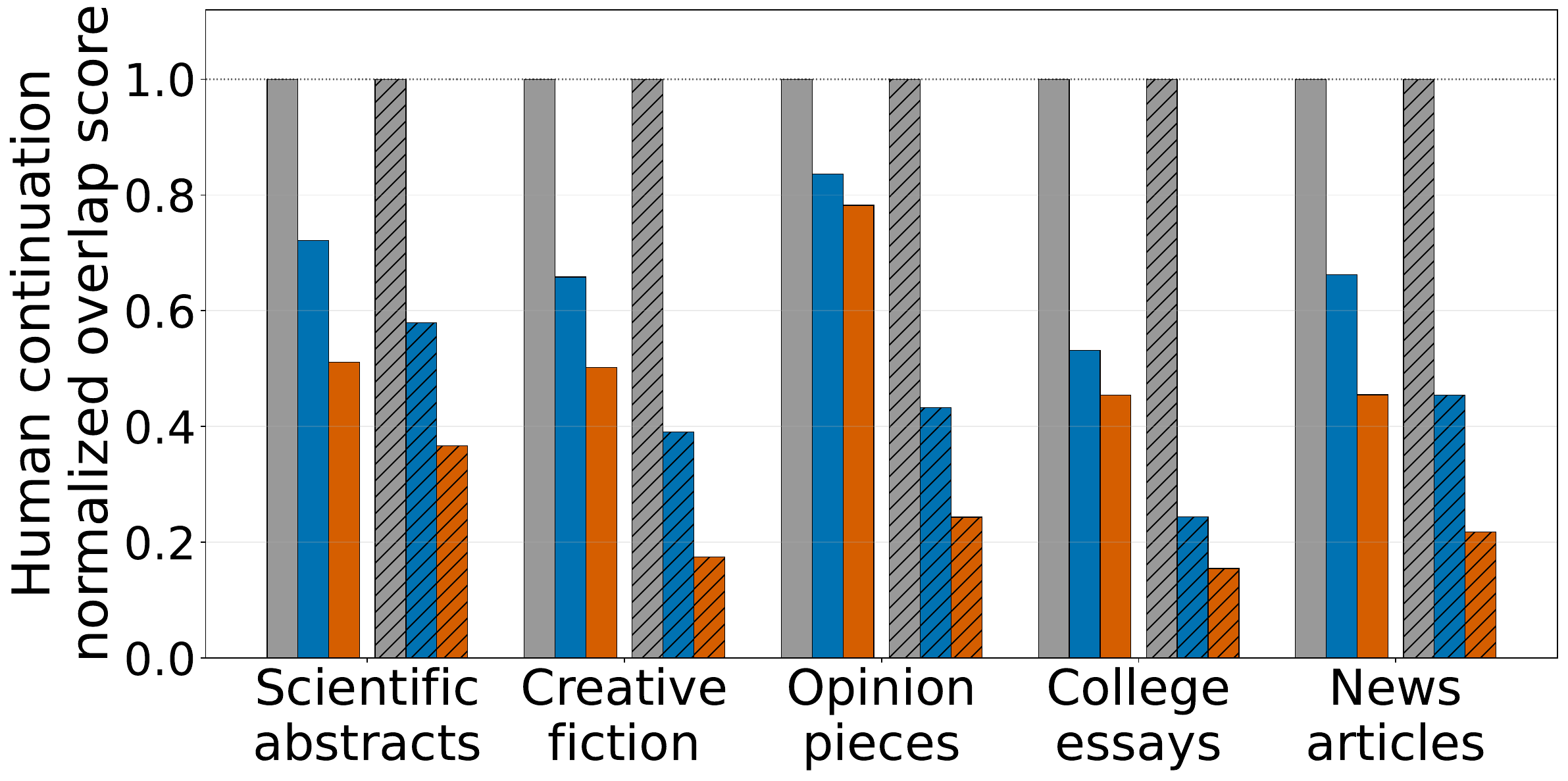}
\caption{
\textbf{Post-training reduces verbatim affinity to human-corpus text.}
For each domain, we compare matched continuations generated from the same human-written prefix:
\legendbox{humangray}{Human},
\legendbox{baseblue}{Base}, and
\legendbox{alignedvermillion}{Aligned}.
Scores are normalized by the human-continuation mean within each domain, so Human $=1.0$ by construction.
Solid bars show overlap density, the mean longest RedPajama+-matched span across word positions; hatched bars show Coverage@10, the fraction of words covered by RedPajama+ matched spans of at least 10 words.
Across domains and metrics, scores follow Human $>$ Base $>$ Aligned, indicating that post-training moves generations further away from source-domain human text.
}
\label{fig:rpj-affinity}
\end{figure}

We instantiate $C$ with a RedPajama+ infini-gram index (details in Appendix~\ref{app:h1_details}) over pre-2023 publicly
available text. The index is a proxy for broad pretraining-style human text,
not evidence of exact membership in any model's training data.
Figure~\ref{fig:rpj-affinity} shows normalized corpus-affinity scores across
scientific abstracts, creative fiction, opinion pieces, college essays, and
news articles, with each domain divided by its human-continuation mean. Both
overlap density and Coverage@10 follow Human $>$ Base $>$ Aligned in every
domain. The drop is largest when the index has close source overlap with the
domain; opinion pieces show a weaker but directionally consistent effect. Thus, aligned generations have lower verbatim affinity to pretraining-style
human text than base generations from the same human-written prefixes,
supporting H1. Full raw scores, normalized scores and metric definitions are reported in
Appendix~\ref{sec:posttraining_appendix}.

\begin{figure}[t]
\centering
\includegraphics[width=\linewidth]{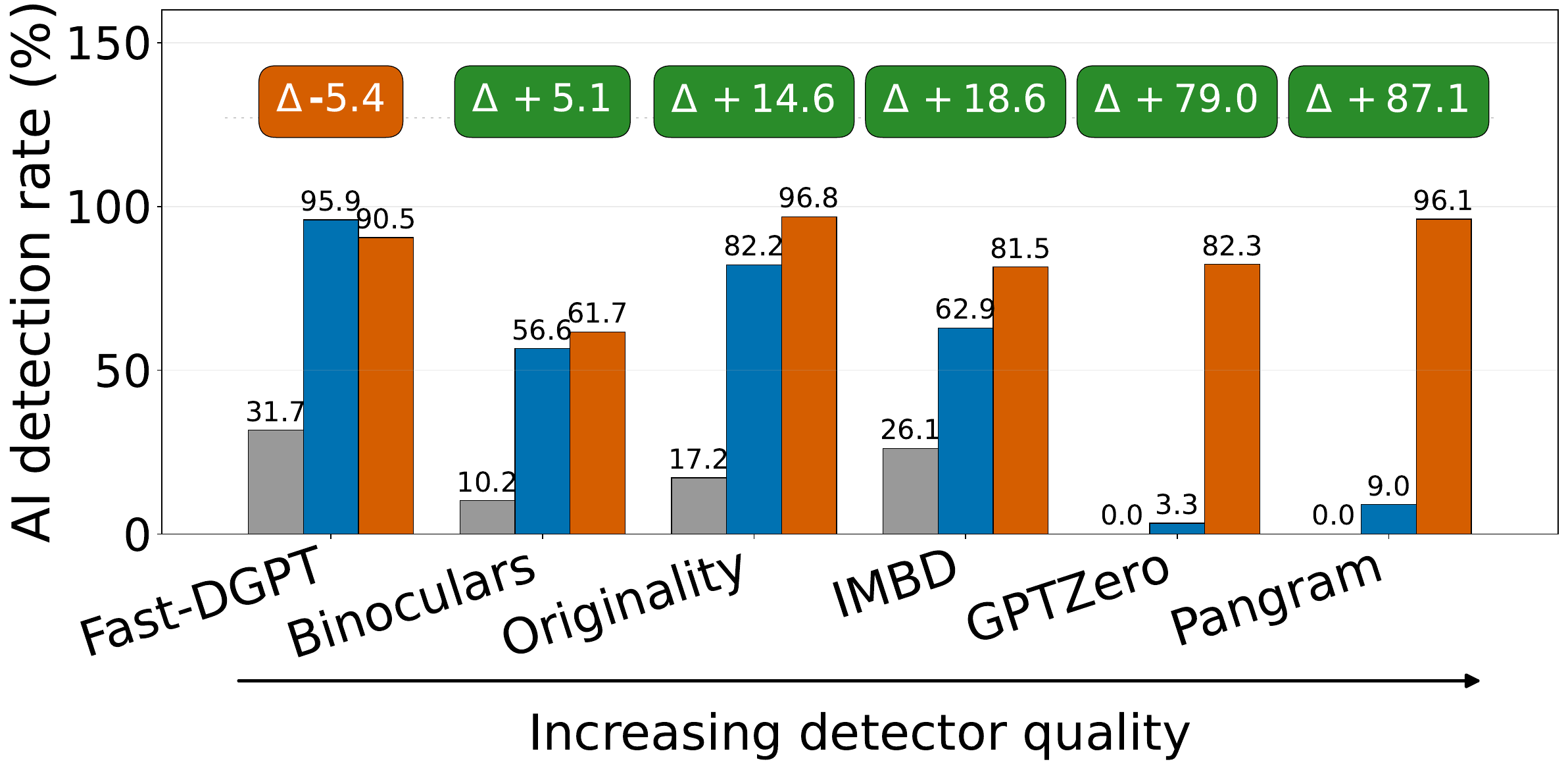}
\caption{
\textbf{Post-training increases AI-detector visibility.}
We report AI-detection rates for
\legendbox{humangray}{Human},
\legendbox{baseblue}{Base}, and
\legendbox{alignedvermillion}{Aligned}
continuations. Base and Aligned outputs are generated from the same human-written prefixes and length-controlled before evaluation. Detectors are ordered from left to right by increasing macro accuracy on our detector-quality evaluation. Higher-quality detectors (Appendix~\ref{sec:ai_detector_quality_eval}) assign much higher AI rates to aligned outputs than to base outputs, supporting H2 that post-training introduces detector-visible regularities.
}
\label{fig:ai-detection-base-vs-aligned}
\end{figure}

\paragraph{H2: Post-training increases AI-detector visibility.}
H2 asks whether the surface shift in H1 is also visible to AI detectors. Let
$d(t)$ denote an AI-detector score or classification rate, where larger values
mean that $t$ is more likely to be classified as AI-generated. Let
$d_{\mathrm{H}}$, $d_{\mathrm{B}}$, and $d_{\mathrm{A}}$ denote the mean detector
score for human, base, and aligned continuations, respectively. H2 predicts
\begin{equation}
d_{\mathrm{H}} < d_{\mathrm{B}} < d_{\mathrm{A}}.
\label{eq:detector-ordering}
\end{equation}

The key comparison is $t_{\mathrm{base}}$ versus $t_{\mathrm{aligned}}$, since
both are generated from the same prefix by models from the same family. If
aligned continuations receive higher detector scores than base continuations,
the increase is not merely a property of machine-generated text. It indicates
that post-training adds detector-visible regularities on top of the base-model
distribution.

We evaluate H2 on same models and domains as H1. To make the test
conservative, we filter base-model continuations for degeneracy and re-rank them
with a Claude-Sonnet-4.6 judge for detecting degeneracy. Thus, each aligned output is
compared against a base output selected to be non-degenerate.

Figure~\ref{fig:ai-detection-base-vs-aligned} reports AI-detection rates for
base and aligned generations across detectors, ordered by accuracy on our
detector-quality evaluation (Appendix~\ref{sec:ai_detector_quality_eval}).
Nearly all detectors flag aligned generations more often than base generations.
The largest gaps are from Pangram and GPTZero: Pangram flags 98.5\% of aligned
outputs versus 9.2\% of base outputs, while GPTZero flags 84.5\% versus 2.9\%.
Fast-DetectGPT is the only exception; which is unreliable, indicated by a higher false-positive rate on
human text.
These results support H2: under matched prefixes, lengths, and model families,
aligned generations are more detector-visible than base generations. The effect
is strongest for higher-accuracy detectors, suggesting that post-training adds
a shared detector-visible style beyond the base-model distribution.

\paragraph{Takeaway: Post-training Leaves a Surface Signature.} H1 and H2 show the same shift from complementary angles. Under matched
human-written prefixes, aligned generations are less similar to
pretraining-style human text and more visible to AI detectors than base
generations. This supports Figure~\ref{fig:pasta-overview}A: post-training
moves generations away from human-corpus-like text and toward detector-visible
assistant-style text.
This surface signature could reflect different token-level statistics, but it does not identify a responsible mechanism. We next test whether part of it is mediated by a locatable residual-stream direction.

\section{PASTA: Post-training Alignment Signature Targeted Ablation}
\label{sec:pasta}

The surface signature in Section~\ref{sec:posttraining} shows
that post-training moves generations toward a more detector-visible
assistant-style regime. We now ask whether part of this shift is mediated by a
locatable internal feature. Our hypothesis is that post-training induces a
residual-stream direction that separates aligned-model behavior from base-model
behavior. If this direction is causally involved in detector-visible style,
then ablating it during decoding should reduce AI-detector visibility.
We call this intervention \pasta: \textbf{P}ost-training \textbf{A}lignment
\textbf{S}ignature \textbf{T}argeted \textbf{A}blation.

\paragraph{Method.}
Let $\mathcal{M}_{\mathrm{base}}$ be a pretrained model and
$\mathcal{M}_{\mathrm{align}}$ its post-trained counterpart. Given a small
calibration set of matched naturalistic prompts, we generate continuations from
both models and record residual-stream activations over generated tokens. For
each layer $\ell$, we compute the mean residual state for each model:
\begin{align}
\mu_{\mathrm{base}}^{\ell}
&= \mathbb{E}_{p,t}\!\left[h_{\mathrm{base}}^{\ell}(p,t)\right], \\
\mu_{\mathrm{align}}^{\ell}
&= \mathbb{E}_{p,t}\!\left[h_{\mathrm{align}}^{\ell}(p,t)\right],
\label{eq:pasta-means}
\end{align}
where $h^{\ell}(p,t)$ is the residual-stream activation at layer $\ell$ for
generated token $t$ under prompt $p$. Their normalized difference defines the
candidate post-training direction:
\begin{equation}
v_{\mathrm{PASTA}}^{\ell}
=
\frac{
\mu_{\mathrm{align}}^{\ell} - \mu_{\mathrm{base}}^{\ell}
}{
\left\|
\mu_{\mathrm{align}}^{\ell} - \mu_{\mathrm{base}}^{\ell}
\right\|_2
}.
\label{eq:pasta-direction}
\end{equation}

To test whether this direction is causally involved in detector-visible style,
we ablate it during aligned-model decoding. For a residual vector $x$ and a
unit direction $v$, \pasta removes the component of $x$ along $v$ with strength $\alpha$:
\begin{equation}
\mathrm{Ablate}_{\alpha}(x;v)
=
x - \alpha (x^\top v)v
\label{eq:pasta-ablation}
\end{equation}
This projection, with default $\alpha=1$, is applied during autoregressive generation at residual-stream write sites of the aligned model.
To choose the most effective $v$, we first perform a detector-guided layer sweep on the calibration set. 
For each
candidate layer $\ell$, we generate ablated continuations using
$v_{\mathrm{PASTA}}^{\ell}$ and score them with a detector $\mathcal{D}$. We
select the layer $\ell^*$ whose direction yields the lowest calibration-set
AI-detection rate, then fix $v_{\mathrm{PASTA}}^{\ell^*}$ for held-out test
prompts. Figure~\ref{fig:pasta-overview} summarizes the pipeline, and
Appendix~\ref{app:pasta-algo} gives full pseudocode.

\paragraph{Experimental Setup.}
We draw prompts from the same five source datasets used in Section~\ref{sec:posttraining}: college essays from \texttt{qwedsacf/ivypanda-essays}, news articles from \texttt{abisee/cnn\_dailymail}, scientific abstracts from \texttt{sentence-transformers/s2orc}, creative fiction from \texttt{euclaise/writingprompts}, and opinion pieces from \texttt{fabiochiu/medium-articles} (Appendix~\ref{app:prefix-extraction}).
For each aligned model, we estimate a single \pasta direction using 25 randomly selected
examples\footnote{Using 100 calibration examples yielded directions similar to 25-example directions with a cosine similarity of $0.92$--$0.96$ across 4 models.}, with five examples from each domain. The direction is selected with
Pangram on the pooled calibration set and then applied to 500 held-out examples,
with 100 examples per domain. Calibration and test prompts are disjoint. Unless
otherwise specified, we use Pangram for detector-guided layer selection because
it is the strongest detector in our detector-quality evaluation.
We evaluate the resulting generations with Pangram, Binoculars, ImBD, and
RAIDAR. Pangram measures performance on the selection detector, while the
remaining detectors test cross-detector transfer. For each model $m$ and
detector $d$, we report the signed change in AI-detection rate after \pasta
ablation:
\[
\Delta^{\pasta}_{m,d}
=
\mathrm{AI}^{\pasta}_{m,d}
-
\mathrm{AI}^{\mathrm{aligned}}_{m,d},
\]
where $\mathrm{AI}^{c}_{m,d}$ is the percentage of generations from condition $c$
flagged as AI by detector $d$ for model $m$. Negative values indicate reduced
detector visibility after \pasta ablation.

\paragraph{Results.}
Table~\ref{tab:pasta-main} shows that \pasta reduces detector visibility across
many aligned models. The largest reductions occur on Pangram, the detector used
for layer selection: 8 of 11 models drop by $73.2$--$98.0$ points. These
include Gemma-2-9B, several Qwen2.5 models from 1.5B to 32B, and
Llama-3.1-8B-Instruct.
The effect also transfers beyond the selection detector. ImBD and RAIDAR
decrease for every model, and Binoculars decreases for 10 of 11 models. This
cross-detector transfer suggests that \pasta is not merely overfitting to
Pangram, but removing a broader detector-visible component of the post-training
signature.
The weakest cases are OLMo-3-7B-Instruct and
Tulu-3-8B-RLVR. 
These models show the smallest Pangram reductions
($\leq 6.0$ points), and Binoculars sometimes increases after ablation. We
interpret these exceptions cautiously because
Appendix~\ref{sec:ai_detector_quality_eval} shows that Binoculars is less
reliable than the strongest detectors in our setting. More broadly, these
failures suggest that the strength and linearity of the post-training signature
vary across model families and post-training recipes.
Overall, these results support the mechanistic hypothesis behind \pasta. For
many aligned models, an aligned-minus-base residual-stream direction captures a
detector-visible component of post-training. 
Ablating this direction at
inference time reduces AI-detector visibility, often transfers across
detectors, and requires only a small calibration set with no parameter updates.

\begin{table}[t]
\centering
\scriptsize
\setlength{\tabcolsep}{2.4pt}
\renewcommand{\arraystretch}{1.0}
\caption{
\textbf{\pasta reduces AI-detector visibility and transfers across detectors.}
The second column reports $\mathrm{AI}^{\mathrm{aligned}}_{m,\mathrm{Pang.}}$; remaining columns report $\Delta^{\pasta}_{m,d}$, where negative values indicate reduced detector visibility. Pangram selects the \pasta-layer; other detectors test transfer. Rows are sorted by $\Delta^{\pasta}_{m,Pang.}$.
}
\label{tab:pasta-main}
\begin{tabular}{@{}lrrrrr@{}}
\toprule
\multirow{2}{*}{\textbf{Model (m)}} 
& \multicolumn{1}{c}{\multirow{2}{*}{\textbf{$\mathrm{AI}^{\mathrm{aligned}}_{m,\mathrm{Pang.}}$}}}
& \multicolumn{4}{c}{\textbf{$\Delta^{\pasta}_{m,d}$} for detector $d$} \\
\cmidrule(lr){3-6}
& 
& \textbf{Pang.}
& \textbf{Bino.}
& \textbf{ImBD}
& \textbf{RAID.} \\
\midrule
Gemma-2-9B        & 100.0 & $\mathbf{-98.0}$ & $-20.8$ & $-45.6$ & $-22.4$ \\
Qwen2.5-32B       & 98.2  & $\mathbf{-95.6}$ & $-58.8$ & $-47.3$ & $-13.1$ \\
Qwen2.5-14B       & 98.8  & $\mathbf{-94.8}$ & $-32.6$ & $-35.8$ & $-14.5$ \\
Qwen2.5-1.5B      & 99.0  & $\mathbf{-89.7}$ & $-10.8$ & $-87.2$ & $-18.9$ \\
Qwen2.5-7B        & 98.6  & $\mathbf{-87.2}$ & $-26.0$ & $-14.4$ & $-12.1$ \\
Qwen2.5-3B        & 97.4  & $\mathbf{-74.9}$ & $-47.0$ & $-40.2$ & $-13.4$ \\
Llama-3.1-8B-Inst.& 99.8  & $\mathbf{-73.2}$ & $-26.4$ & $-24.4$ & $-13.4$ \\
Mistral-7B-v0.3   & 99.0  & $\mathbf{-36.4}$ & $-42.9$ & $-7.0$  & $-4.1$  \\
Qwen2.5-0.5B      & 99.2  & $\mathbf{-9.8}$  & $-3.8$  & $-29.4$ & $-10.7$ \\
Tulu-3-8B-RLVR    & 99.0  & $\mathbf{-6.0}$  & $+8.5$  & $-0.4$  & $-2.9$  \\
OLMo-3-7B-Inst.   & 95.6  & $\mathbf{-5.8}$  & $+12.2$ & $-2.4$  & $-2.1$  \\
\bottomrule
\end{tabular}
\end{table}

\section{Analyzing \pasta: Specificity, Quality, and Robustness}
\label{sec:analysis}

We ask why \pasta works and when its effects hold. We test whether reduced AI-detectability is specific
to the \pasta direction rather than a generic effect of residual-stream ablation. We then assess generation quality and robustness across ablation strengths, domains, and detectors. Together, these analyses show that \pasta is direction-specific, preserves high-quality generation, and remains robust across settings.

\subsection{Specificity of the \pasta's Direction}
\label{sec:analysis_random_direction}

We test whether \pasta's reduction in AI-detectability is specific to its learned direction, rather than a generic consequence of ablating residual-stream directions. 
For four aligned models (Gemma2-9B, Llama3.1-8B, Qwen2.5-1.5B, and Qwen2.5-7B), we sample five mutually orthogonal unit random directions $\{v_k\}_{k=1}^{5}$, with $v_k \in \mathbb{R}^d$, each also orthogonal to \pasta's direction.
Each random ablation uses the same residual-stream hook footprint and decoding seed as \pasta. We evaluate on a held-out set of $125$ prompts per model ($5$ domains $\times$ $25$ prompts) using six detectors: Pangram, GPTZero, Binoculars, ImBD, Originality, and Fast-DetectGPT. 
For each model $m$ and detector $d$, we report the best random-direction change,
\[
\Delta^{\mathrm{rand}}_{m,d}
=
\min_{k \in \{1,\dots,5\}}
\left(\mathrm{AI}^{v_k\text{-ablated}}_{m,d}
-
\mathrm{AI}^{\mathrm{aligned}}_{m,d}\right),
\]
where $\mathrm{AI}^{c}_{m,d}$ is the percentage of generations from condition $c$
flagged as AI by detector $d$ for model $m$. Negative values indicate reduced
detector visibility.

\paragraph{Results.}
Random directions rarely reproduce \pasta's effect. As shown in Table~\ref{tab:random-direction-ablation}, \pasta gives the larger reduction for $22$ of $24$ model--detector pairs, with the largest gaps on the strongest detectors (Appendix~\ref{sec:ai_detector_quality_eval}). For Pangram, the best random direction changes detection by at most $-1.6$ points, compared with $-69.8$ to $-99.2$ points for \pasta. For GPTZero, random directions reach only $-1.2$ to $-5.9$ points, versus $-43.5$ to $-71.5$ points for \pasta. The two exceptions occur on lower-accuracy detector settings, Binoculars on Qwen2.5-1.5B and Fast-DetectGPT on Gemma-2-9B, and are therefore likely detector-specific noise. Overall, this control suggests that \pasta's effect is direction-specific rather than a generic artifact of residual-stream ablation.

\begin{table}[t]
\centering
\scriptsize
\setlength{\tabcolsep}{2.5pt}
\renewcommand{\arraystretch}{1.05}
\caption{
\textbf{Random ablations do not reproduce \pasta's effect.}
Cells report $\Delta^{\mathrm{rand}}_{m,d}/\Delta^{\pasta}_{m,d}$, where each $\Delta$ is a post-ablation change in AI-detection rate. Random ablations use the best of five mutually orthogonal directions, each also orthogonal to the selected \pasta direction. Negative values indicate reduced detector visibility; bold marks the larger reduction. Rows are ordered by detector quality.
}
\label{tab:random-direction-ablation}
\resizebox{\columnwidth}{!}{
\begin{tabular}{@{}lcccc@{}}
\toprule
\textbf{Detector}
& \makecell{\textbf{Gemma2-9B}}
& \makecell{\textbf{Llama3.1-8B}}
& \makecell{\textbf{Qwen2.5-1.5B}}
& \makecell{\textbf{Qwen2.5-7B}} \\
\midrule
Pangram
& $-1.6/\mathbf{-99.2}$
& $0.0/\mathbf{-69.8}$
& $-0.8/\mathbf{-90.4}$
& $-0.8/\mathbf{-84.0}$ \\
GPTZero
& $-1.2/\mathbf{-66.0}$
& $-1.6/\mathbf{-55.1}$
& $-5.9/\mathbf{-43.5}$
& $-3.7/\mathbf{-71.5}$ \\
ImBD
& $-2.4/\mathbf{-36.0}$
& $0.0/\mathbf{-19.2}$
& $-7.2/\mathbf{-89.6}$
& $-0.8/\mathbf{-13.6}$ \\
Originality
& $0.0/\mathbf{-26.9}$
& $-1.6/\mathbf{-19.4}$
& $-1.6/\mathbf{-82.7}$
& $+0.7/\mathbf{-27.3}$ \\
Bino
& $-1.6/\mathbf{-20.0}$
& $-1.6/\mathbf{-34.4}$
& $\mathbf{-8.0}/-7.2$
& $-0.8/\mathbf{-32.0}$ \\
FastDGPT
& $\mathbf{-1.1}/+1.2$
& $-0.3/\mathbf{-6.6}$
& $-9.1/\mathbf{-46.6}$
& $-0.1/\mathbf{-5.5}$ \\
\bottomrule
\end{tabular}
}
\end{table}

\subsection{\pasta Preserves Generation Quality}
\label{sec:qualitative_analysis}

We test whether reducing detector-visible style affects generation quality.
For each human-sourced prefix, we compare a triplet of continuations: base, aligned, and \pasta-ablated. Our evaluation contains $50$ such triplets, drawn from $5$ aligned models (Qwen2.5-7B, Llama3.1-8B, Qwen2.5-32B, Tulu3-8B, OLMo3-7B) and $5$ domains.
We use Claude Sonnet 4.6 as a pairwise LLM judge. For each prefix, we compare all possible pairs of continuations from a triplet of (base, aligned, and \pasta-ablated); 
the judge assigns rubric-based scalar scores to both continuations and selects the preferred one. To control for position bias, we evaluate each pair in both presentation orders.
We evaluate two metrics: (a) \textbf{relevance}, covering instruction following, topical or genre adherence; and (b) \textbf{fluency}, covering grammaticality, coherence, and readability.
For each metric, let $\mathcal{N}_{X,Y}$ be the set of pairwise comparisons between continuation types $X$ and $Y$. We define
$\mathrm{WR}(X/Y)=\big(W_{X/Y}+0.5T_{X/Y}\big)/|\mathcal{N}_{X,Y}|$,
where $W_{X/Y}$ is the number of wins for $X$ and $T_{X/Y}$ is the number of ties; higher $\mathrm{WR}(X/Y)$ indicates that $X$ is preferred more often under that criterion. We also report
$\Delta(X/Y)=\mathbb{E}[\mathrm{score}(X)]-\mathbb{E}[\mathrm{score}(Y)]$,
with positive values indicating higher average rubric scores for $X$. We include more details on our evaluation setup and LLM-judge in Appendix~\ref{app:judge_prompts}\&~\ref{app:qual_metrics}.
We additionally conducted a small-scale human evaluation to independently validate the judge LLM's assessments, and report the methodology and results in Appendix~\ref{app:human_evaluation}.

\begin{figure}[t]
\centering
\includegraphics[width=\linewidth]{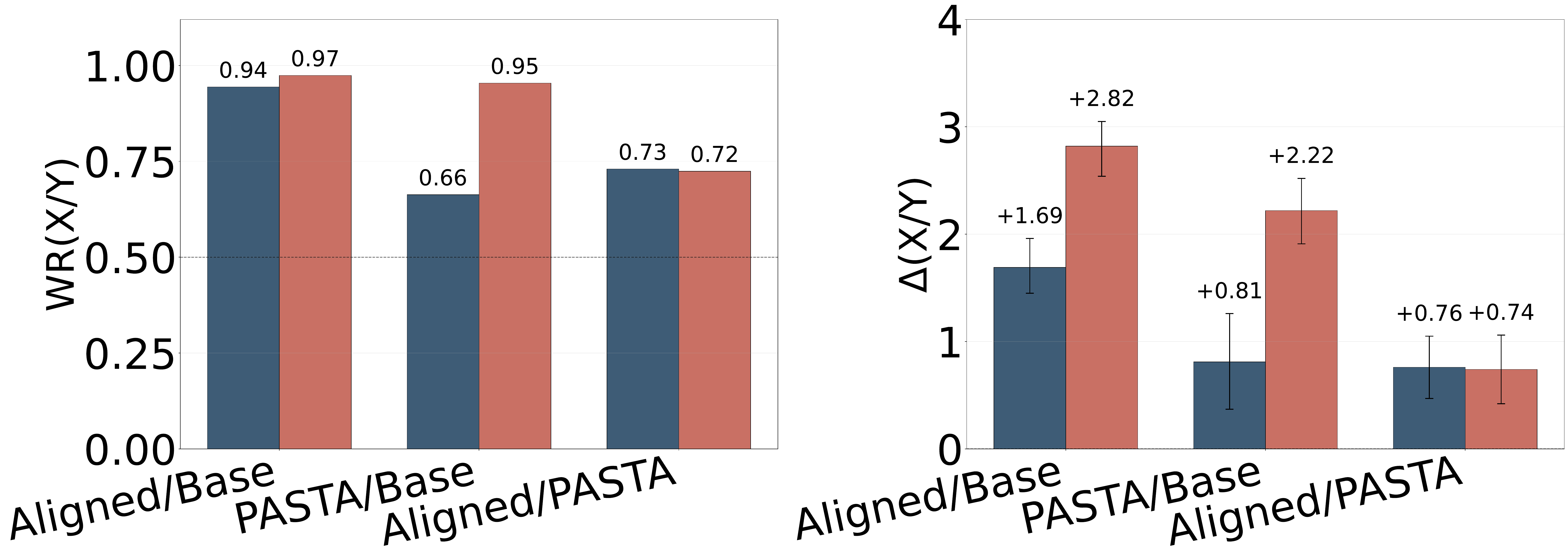}
\caption{
\textbf{\pasta preserves much of the aligned model's fluency and relevance.}
Pairwise Claude Sonnet 4.6 judgments compare aligned, base, and \pasta-ablated continuations for
\legendbox{fluencycolor}{Fluency} and
\legendbox{relevancecolor}{Relevance}.
Left: tie-adjusted win rates $\mathrm{WR}(X/Y)$; right: mean rubric-score differences
$\Delta(X/Y)$ with 95\% cluster-bootstrap confidence intervals.
Higher values favor the first system in each pair. \pasta remains substantially better than base and only modestly below aligned.
}
\label{fig:fluency-relevance-pairwise-winrate}
\end{figure}

\paragraph{Results.}
Figure~\ref{fig:fluency-relevance-pairwise-winrate} shows a quality hierarchy. 
Aligned generations strongly outperform base generations in fluency 
($\mathrm{WR}=0.944$, $\Delta=+1.69$) and relevance 
($\mathrm{WR}=0.974$, $\Delta=+2.82$), validating the judge setup. 
\pasta remains substantially better than base, with gains in fluency 
($\mathrm{WR}=0.663$, $\Delta=+0.81$) and especially relevance 
($\mathrm{WR}=0.954$, $\Delta=+2.22$). 
Although aligned generations outperform \pasta, the remaining gap is much smaller than the aligned--base gap: $+0.76$ vs.\ $+1.69$ in fluency and $+0.74$ vs.\ $+2.82$ in relevance. 
We additionally validate the judge's assessment with a human evaluation of $60$ pairwise comparisons across aligned, base, and \pasta generations. 
The human evaluation agrees with the LLM judge in the direction of all six pairwise win-rate comparisons. Moreover, the human-judged gap between aligned and \pasta remains smaller than the gap between aligned and base generations in both fluency ($+0.50$ vs.\ $+0.86$) and relevance ($+0.00$ vs.\ $+1.00$).
Thus, \pasta preserves a large fraction of the quality gains from post-training while removing a detector-visible direction, rather than reverting aligned models to base behavior.

\paragraph{Qualitative Patterns.}
To interpret the source of the aligned--\pasta gap, we inspect matched aligned and \pasta generations in Appendix~\ref{app:pasta_quality_examples}. The examples suggest that many aligned--\pasta differences are stylistic rather than topical: \pasta generations often become more direct, drop rhetorical or Markdown-like structure, and reduce polished alignment-style phrasing, while preserving the main topic or thesis. In mild tradeoff cases, \pasta loses some detail, format adherence, or argumentative polish; in failure cases, it can also degrade fluency and coherence. These patterns are consistent with the judge results: \pasta preserves much of the aligned model's relevance and fluency, but the removed direction also carries some style- and quality-relevant information.

\subsection{\pasta Robustness Across Settings}
\label{sec:pasta-robustness}

We test whether \pasta depends on a narrowly tuned setup. We consider three axes of robustness: ablation strength, domain specialization, and detector choice for direction selection. We also provide detailed post-training stage, domain, detector, and direction-similarity breakdowns in Appendix~\ref{app:pasta-robustness}.

\paragraph{Ablation strength.}
We vary intervention strength by scaling the selected direction as $\alpha * v_{\pasta}$, where $\alpha=0$ is the aligned model and $\alpha=1$ is the default \pasta setting. Figure~\ref{fig:scaled-ablation-genquality-frontier} shows that the detector effect appears before the default setting: even $\alpha=0.25$ produces a large drop in AI-detection rate, and moderate values remain well below the aligned baseline. Thus, \pasta is not brittle to a single scale. However, lower detectability is not free. Larger $\alpha$ values can further lower detector scores, but increasingly degrade fluency and prompt relevance. Scaling \pasta therefore exposes a detectability--quality frontier: moderate ablation removes much of the detector-visible signature, while aggressive ablation can reduce detection partly by damaging generation quality.

\paragraph{Domain specialization.}
We next ask whether \pasta must be estimated separately for each domain. For Qwen2.5-7B-Instruct and Llama3.1-8B-Instruct, we compare a domain-agnostic direction against directions extracted from each target domain. Figure~\ref{fig:in-domain-pareto} shows no consistent Pareto gain from domain specialization. If in-domain \pasta were reliably better, arrows would move toward lower detection and higher quality. Instead, they scatter. Using Pangram and GPTZero, domain-agnostic \pasta gives lower mean detection in $6/10$ model--domain cells, compared with $3/10$ for domain-specific \pasta and one tie; quality differences are small. This suggests that a domain-agnostic direction captures a broadly useful  signature rather than a domain-specific artifact.

\paragraph{Direction-Selecting Detector.}
Finally, we test whether the detector used for direction selection changes the ablation. Detector-specific extractors are strongest against their matching evaluators: matched extractor--evaluator pairs give the lowest detection in Appendix~\ref{sec:detector_swap}. This indicates that detectors emphasize partly different surface cues. At the same time, the Pangram-selected direction transfers to other detectors in the main results, so \pasta is not limited to the detector used for selection. We use Pangram as the default extractor because it is strongest in our detector-quality evaluation, and treat detector-specific extraction as an expected specialization.

\section{Discussion}
\label{sec:discussion}

Our results suggest that AI-like style is not merely an output-level artifact. 
Aligned generations become more regular, polished, structured, and separable from human text, and part of this shift can be localized to a residual-stream direction. 
Other factors also shape generation style, including pretraining data, synthetic-text contamination, decoding choices, model scale, and prompting conventions. 
However, post-training is especially important because it deliberately shapes model behavior toward helpfulness, safety, and instruction following. 
Our results show that this shaping leaves a measurable representational footprint: \pasta turns ``AI-like writing'' into an object that can be estimated, removed, and causally tested.

This perspective also reframes AI detection. 
Alignment improves usability, but it can narrow generations toward a recognizable style: polished, hedged, affirming, and predictably structured. 
Detectors may exploit this narrowness, so detector visibility is not only a signal of machine authorship; it can also reflect post-training-induced stylistic regularity. 
The goal is not to remove alignment, but to achieve better alignment: methods that preserve safety and instruction following without collapsing stylistic diversity. 
In this view, \pasta is not a detector-evasion method; it is a tool for studying how post-training shapes style representations.

\section{Conclusion}
\label{sec:conclusion}

We studied whether the recognizable AI-like writing style of aligned language models is only an output-level pattern or instead reflects a localized representational change from post-training. 
Under matched human-source prefixes, aligned generations move away from human-corpus text and become more visible to AI detectors than base generations. 
We introduced \pasta, a training-free intervention that estimates an aligned-minus-base residual direction and projects it out during inference. 
Across models and detectors, \pasta reduces detector visibility, transfers beyond the detector used for direction selection, is not reproduced by random directions, and preserves much of the aligned model's fluency and relevance. 
Together, these results show that post-training leaves a measurable representation-level stylistic signature, suggesting a path toward post-training methods that preserve helpfulness and safety without collapsing generations into a narrow, uniformly recognizable style.

\section*{Limitations}
\label{sec:limitations}

\paragraph{Proxy-Based Definition of AI-Like Style.}
We operationalize AI-like style through two proxies: human-corpus affinity and AI-detector visibility. These probes are useful because they expose stylistic separability, but they do not completely define human-likeness or authorship. Lower corpus affinity means reduced verbatim proximity to a reference corpus, not worse naturalness. Lower detector visibility means reduced detector-visible regularity, not proof of human authorship or robust detector evasion. Our claims, therefore, concern a measurable, detector-visible component of AI-like style, not the full space of differences between human and model-generated writing.

\paragraph{Prompt Format Confound in Base-Aligned Comparisons.}
When comparing the generations of base and aligned models, the base models receive raw-text continuation prompts, whereas aligned models get their model-appropriate chat templates with a "Begin with:" anchor. Therefore, part of the observed base-aligned shift may be due to the difference in prompt format rather than post-training alone. We nonetheless retain the model-appropriate prompts because forcing aligned models to continue from base-model prompts is against its intended practical usage. We found that forcibly doing so, substantially degrades generation quality - as a control for H2, we evaluate three representative model pairs using the same raw-text prompt and still find large increments in AI-detection rate of aligned models ($+50.7$, $+30.8$, and $+89.9$ percentage points for Llama-3.1-8B, Qwen2.5-7B, and Gemma-2-9B, respectively). Prompt format thus accounts for part of the observed shift, but a substantial gap remains, and H2 continues to hold.

\paragraph{Limited Attribution Within the Post-Training Stack.}
Our base--aligned comparisons localize the shift to the released post-training stack, but not to a specific ingredient inside that stack. The signature may arise from instruction data, preference objectives, safety policies, chat templates, or response-formatting conventions, and related AI-like style can also be shaped by pretraining data, synthetic text, decoding choices, and model scale. Stronger causal attribution would require controlled post-training pipelines that vary these factors independently while holding the base model fixed.

\paragraph{Partial Mechanistic Account.}
PASTA provides a causal handle on the post-training signature, but not a complete mechanistic account. 
Its success shows that one detector-visible component of post-training is captured by a linearly removable residual-stream direction. 
This does not imply that AI-like style is a single-direction phenomenon. 
Other components may be distributed across layers, pathways, or nonlinear interactions. 
Nor is the identified direction purely stylistic: stronger ablation can reduce detector visibility while also harming fluency, relevance, format adherence, or coherence. 
Thus, \pasta exposes one important representational component of post-training style, but does not fully explain how that style is produced.

\paragraph{Evaluation Scope and Generality.}
Finally, we use pairwise LLM-judge comparisons and qualitative human inspection to assess fluency, relevance, factual plausibility, argument quality, and narrative quality. 
While this gives scalable and interpretable evidence, a direct human study would better test how readers perceive \pasta generations in practice, including usefulness, naturalness, prompt faithfulness, and stylistic diversity in longer writing tasks. 
Our experiments also focus on English text domains and open-weight models with base-aligned counterparts, which helps isolate post-training effects but leaves open whether the same signatures hold for multilingual generation, code, dialogue, tool use, multimodal outputs, closed-source systems, or different post-training recipes. 
Testing these settings would clarify which aspects of the alignment signature are broadly shared and which are specific to our domains and detectors.

\section*{Ethics Statement}
\label{sec:ethics}

\paragraph{Misuse Potential and Detector Evasion.}
This work studies an inference-time intervention that can reduce the detectability of aligned-model generations by AI-text detectors. Although detector scores are used here only as probes of stylistic separability, the same intervention could be misused to make machine-generated text harder to detect. We therefore present \pasta as a tool for studying post-training-induced stylistic signatures, not as a method for detector evasion. More broadly, our results suggest that the detectors we study may rely partly on alignment-induced stylistic artifacts, highlighting the need for more robust detection methods and post-training procedures that preserve alignment without imposing overly narrow stylistic signatures.

\paragraph{Use of AI Assistants.}
Claude was used as an auxiliary tool in this work, not as a source of research ideas, methods, analyses, interpretations, or claims. Its role was limited to locating and summarizing potentially relevant papers, supporting code implementation and debugging, and polishing the grammar, clarity, and concision of author-written text. All literature judgments, conceptual framing, methodological choices, experimental designs, analyses, claims, and conclusions were made by the authors. All AI-assisted text, code, and other outputs were independently reviewed, edited, and accepted only after author verification, and the authors take full responsibility for the correctness of the final manuscript.

\section*{Acknowledgements}
\label{sec:acknowledgements}

We thank Ari Holtzman for early discussions and for directing us to related ideas on alignment-induced stylistic signatures in language models. We thank Kartik Sharma for thoughtful draft feedback. We gratefully acknowledge Pangram Labs, Originality.ai, and GPTZero for providing detector credits that enabled the empirical evaluations in this work. This work used the DeltaAI system at the National Center for Supercomputing Applications through ACCESS allocation CIS251207, supported by award OAC 2320345. ACCESS is supported by NSF grants \#2138259, \#2138286, \#2138307, \#2137603, and \#2138296.

\bibliography{custom}

\appendix

\newpage
\onecolumn

\section*{Appendix Contents}
\label{app:contents}

\begingroup
\setlength{\parindent}{0pt}
\setlength{\parskip}{0.55em}

\newcommand{\appsectionentry}[2]{%
  \noindent
  \begin{tabular*}{\textwidth}{@{}p{2.2em}@{\hspace{0.8em}}p{0.76\textwidth}@{\extracolsep{\fill}}r@{}}
    \textbf{\ref*{#1}} &
    \hyperref[#1]{\textbf{#2}} &
    \textbf{\pageref*{#1}}
  \end{tabular*}\par
}

\newcommand{\appsubsectionentry}[2]{%
  \noindent
  \begin{tabular*}{\textwidth}{@{}p{4.2em}@{\hspace{0.8em}}p{0.70\textwidth}@{\extracolsep{\fill}}r@{}}
    \hspace{2.0em}\ref*{#1} &
    \hyperref[#1]{#2}\dotfill &
    \pageref*{#1}
  \end{tabular*}\par
}

\appsectionentry{app:prefix-extraction}{Dataset Details}
\appsubsectionentry{app:domain_selection}{Domain selection criteria}

\appsectionentry{app:triplet-gen-settings}{Triplet Generation Details}
\appsubsectionentry{app:gen-settings}{Generation Settings}
\appsubsectionentry{app:prompts}{Prompts}

\appsectionentry{app:h1_details}{Additional Details on H1}
\appsubsectionentry{sec:posttraining_appendix}{Additional Corpus-Affinity Results}
\appsubsectionentry{app:ngram-index}{Pretraining-style reference index}
\appsubsectionentry{app:ngram-index:sources}{Indexed sources}
\appsubsectionentry{app:ngram-index:impl}{Index implementation}
\appsubsectionentry{app:ngram-index:preprocessing}{Source-side preprocessing}
\appsubsectionentry{app:ngram-index:scoring}{Query-time scoring}
\appsubsectionentry{app:ngram-index:normalization}{Surface-form normalization}

\appsectionentry{app:h2_details}{Additional Details on H2}
\appsubsectionentry{app:detector-details}{Detector Details}
\appsubsectionentry{sec:ai_detector_quality_eval}{Evaluating AI-detector quality}
\appsubsectionentry{app:h2-breakdown-domain}{H2 Results Breakdown by Domain}
\appsubsectionentry{app:h2-breakdown-models}{H2 Results Breakdown by Models}

\appsectionentry{app:pasta_details}{\pasta Details}
\appsubsectionentry{app:pasta-algo}{Algorithm}

\appsectionentry{app:qual_details}{Additional Details on Quality of \pasta}
\appsubsectionentry{app:judge_prompts}{Prompts for LLM-Judge}
\appsubsectionentry{app:qual_metrics}{Quality Evaluation Metrics}
\appsubsectionentry{app:human_evaluation}{Human Evaluation}
\appsubsectionentry{app:pasta_quality_examples}{Qualitative Patterns in \pasta Generations}

\appsectionentry{app:pasta-robustness}{Additional Results on \pasta's Robustness Across Settings}
\appsubsectionentry{sec:characterize_pasta_direction}{Scaling \pasta Reveals a Detectability--Quality Frontier}
\appsubsectionentry{sec:domain_specifc_agnostic_analysis}{Domain-specific \pasta Does Not Reliably Improve the Detection-Quality Tradeoff}
\appsubsectionentry{sec:detector_swap}{Detector-specific extractors are most effective against their matching evaluators.}
\appsubsectionentry{sec:posttraining_stages}{Contribution of different posttraining stages in AI-sounding generations}
\appsubsectionentry{app:pasta-direction-similarity}{Similarity of \pasta directions}

\endgroup

\newpage

\twocolumn
\section{Dataset Details}
\label{app:prefix-extraction}

This appendix describes the source datasets and the per-domain
methodology used to obtain the 500 text prefixes (5 domains $\times$
100 prefixes) that seed every Base/Aligned generation reported in
Section~\ref{sec:posttraining}. The 100 prefixes within each domain
are sampled deterministically with a fixed seed and subset of these same
500 records are reused across all model pairs.

\subsection{Domain selection criteria}
\label{app:domain_selection}

We chose 5 domains spanning academic, journalistic, scientific,
narrative, and persuasive registers (Table~\ref{tab:domain-datasets}).
Each domain satisfies three constraints: (i) a publicly available,
HuggingFace-resolvable corpus, (ii) entries long enough to sustain a
$\geq 250$-word continuation (the recommended minimum context for
commercial AI detectors), and (iii) where possible, a pre-ChatGPT
provenance to reduce contamination risk in the source human text.

\begin{table}[h]
\centering
\footnotesize
\setlength{\tabcolsep}{3pt}
\caption{Source datasets per domain. \textit{Field} is the dataset
column from which the prefix is extracted.}
\label{tab:domain-datasets}
\begin{tabular}{@{}p{0.28\linewidth}p{0.46\linewidth}p{0.18\linewidth}@{}}
\toprule
Domain & HuggingFace dataset & Field \\
\midrule
College Essays \cite{ivypanda}
  & \texttt{qwedsacf/}\newline\texttt{ivypanda-essays}
  & \texttt{TEXT} \\
\addlinespace
News Articles \cite{see2017get}
  & \texttt{abisee/}\newline\texttt{cnn\_dailymail} (\texttt{3.0.0})
  & \texttt{article} \\
\addlinespace
Scientific Abstracts \cite{lo2020s2orc}
  & \texttt{sentence-transformers/}\newline\texttt{s2orc}
  & \texttt{abstract} \\
\addlinespace
Creative Fiction \cite{fan2018hierarchical}
  & \texttt{euclaise/}\newline\texttt{writingprompts}
  & \texttt{story} \\
\addlinespace
Opinion Pieces \cite{mediumarticles}
  & \texttt{fabiochiu/}\newline\texttt{medium-articles}
  & \texttt{text} \\
\bottomrule
\end{tabular}
\end{table}

\paragraph{Artifact Licenses and Terms.}
We use publicly available datasets, models, detectors, and tools only for research evaluation, and cite the corresponding creators and sources where they are introduced. We do not redistribute the underlying datasets, model weights, or detector services as part of this submission. Users of any released code should obtain these artifacts from their original sources and comply with the corresponding licenses and terms of use.

\paragraph{Artifact Intended Use.}
We use existing datasets, model weights, detectors, and tools for research analysis and evaluation, and access them through their original distribution channels or service interfaces. To the extent that intended uses or access conditions are specified by the original providers, our use is intended to be consistent with those conditions: we do not redistribute the underlying datasets, model weights, or detector services, and any released code or derived artifacts are intended for research on post-training, model analysis, and detector reliability. They should not be used to violate the licenses or terms of the original artifacts, or to support deceptive use of machine-generated text.

\paragraph{Personally Identifying and Offensive Content.}
We use existing publicly available text datasets and do not collect private communications or new human-subject data. Because some source domains, especially news, Reddit fiction, essays, and opinion articles, may contain names of public figures, authors, or other identifying references, and may also contain sensitive or offensive language, we treat the source text as potentially containing such content. We do not use author identities or user identifiers as model inputs or evaluation labels, and we report only aggregate results. We also do not redistribute the underlying source datasets, model weights, or detector services; any released derived data excludes unnecessary source metadata such as author names, URLs, and user identifiers where they are not needed for reproducibility.

\paragraph{College Essays} (\texttt{qwedsacf/ivypanda-essays}, train split). 
The \texttt{TEXT} field contains the full body of a student-authored IvyPanda essay. We take the first $\sim$100 \texttt{cl100k\_base} tokens of \texttt{TEXT} as the prefix and the remainder as the human continuation baseline. The essay topic is recovered from the source URL slug. All of the dataset comes prior to February 3rd, 2023, which is when it was uploaded on HuggingFace; we further verified the \texttt{datePublished} metadata of 95 of 100 sources spans May~2018 to November~2022, and 92/95 were published before ChatGPT's launch. We were unable to verify the metadata of the remaining 5 sources due to their URLs returning \texttt{404}.

\paragraph{News Articles} (\texttt{abisee/cnn\_dailymail}, config \texttt{3.0.0}, train split). 
The non-anonymized \texttt{article} field contains a CNN or Daily~Mail news article. The first $\sim$75 \texttt{cl100k\_base} tokens of \texttt{article} form the prefix; the journalist-written \texttt{highlights} field is preserved as the natural-language summary of the article. We choose 75 tokens (rather than 100) because news leads are informationally dense and a shorter prefix is sufficient to fix the article's framing. The CNN articles in this dataset were written between April~2007 and April~2015.

\paragraph{Scientific Abstracts} (\texttt{sentence-transformers/s2orc}, train split).
The \texttt{abstract} field provides a paper abstract from the Semantic~Scholar Open Research Corpus. The first $\sim$80 \texttt{cl100k\_base} tokens of \texttt{abstract} are taken as the prefix; the paper \texttt{title} is preserved in \texttt{source\_meta}. 
The S2ORC corpus filters out papers published before 1969 and was released in July~2020 \cite{lo2020s2orc}, so the abstract publication window is bounded by roughly 1969-2020; per-record year is not exposed in the \texttt{sentence-transformers/s2orc} subset.

\paragraph{Creative Fiction} (\texttt{euclaise/writingprompts}, train split). 
The \texttt{story} field contains a Reddit \textit{r/WritingPrompts} reply to a prompt thread. The first $\sim$100 \texttt{cl100k\_base} tokens of \texttt{story} form the prefix. The original \texttt{[WP]}-style thread title is preserved as the original prompt; we strip the leading bracketed tag (\texttt{[WP]}, \texttt{[TT]}, \texttt{[EU]}, etc.) before storing. \citet{fan2018hierarchical} report scraping ``three years of prompts and their associated stories'' via the Reddit API for an EMNLP 2018 submission, so the records span roughly 2015 to early~2018; per-record post date is not preserved in the HuggingFace release.

\paragraph{Opinion Pieces} (\texttt{fabiochiu/medium-articles}, train split). 
The dataset contains Medium articles with metadata (\texttt{title}, \texttt{tags}, \texttt{authors}, \texttt{timestamp}, \texttt{url}). We filter to articles whose \texttt{tags} include at least one of \{\textit{Self~Improvement}, \textit{Life}, \textit{Entrepreneurship}, \textit{Creativity}\} and whose \texttt{text} body is at least 800 characters, then take the first $\sim$100 \texttt{cl100k\_base} tokens of \texttt{text} as the prefix. The article \texttt{title} is preserved as the topic descriptor. Based on the \texttt{timestamp} fields, the 100 records that we selected spans December~2015 to October~2021.

For each domain we shuffle the candidate pool with a fixed seed
(\texttt{seed=42}) and retain the first 100 entries that pass the
domain-specific filters above. Each retained record has fields
\{\texttt{id}, \texttt{prefix}, \texttt{instruction},
\texttt{human\_continuation}, \texttt{prefix\_char\_len},
\texttt{prefix\_token\_count}, \texttt{source\_meta}\}, where
\texttt{instruction} is the natural-language description derived from
the dataset's existing fields (e.g., \texttt{highlights} for news,
thread title for fiction) and \texttt{human\_continuation} is the
human-authored continuation of \texttt{prefix} from the same source
record (used elsewhere as a human baseline).

\section{Triplet Generation Details}
\label{app:triplet-gen-settings}

\paragraph{Compute Budget and Infrastructure.}
We ran the open-weight model generation, activation extraction, layer-sweep, and ablation experiments primarily on NVIDIA H100 GPUs. A small number of open-source detector runs were run on NVIDIA A40 GPUs. Model weights were loaded in bfloat16 with no quantization. For a representative 7--9B model in the main PASTA experiment, the computation consists of base/aligned continuation generation on 500 prompts, calibration-time activation extraction on 25 prompts, a layer sweep over calibration prompts, and held-out PASTA generation on 500 prompts. This representative run required approximately 25 H100 GPU-hours. Scaling this estimate by model size and aggregating over the 11 main models in Table~\ref{tab:models}, the main model-generation and intervention experiments required approximately 370 GPU-hours. Additional analyses, including random-direction controls, ablation-strength sweeps, domain-specialized directions, detector-choice analyses, post-training-stage analyses, and small reruns for qualitative evaluation, required approximately 280 GPU-hours. Thus, the total compute budget for model-generation and intervention experiments was approximately 650 GPU-hours. Running open-source AI-text detectors, including ImBD, Fast-DetectGPT, Binoculars, and RAIDAR, required an additional approximately 120 GPU-hours across H100 and A40 GPUs. The total local GPU budget was therefore approximately 750 GPU-hours.

\subsection{Generation Settings}
\label{app:gen-settings}
All continuations are produced with nucleus sampling using identical decoding
hyperparameters for the base and aligned models: temperature $T=1.0$,
top-$p=0.95$, and \texttt{max\_new\_tokens}$=400$. Top-$k$ filtering is
disabled (top-$p$ only), and we apply no repetition or length penalty
(\texttt{num\_beams}$=1$, \texttt{repetition\_penalty}$=1.0$). All model
weights are loaded in \texttt{bfloat16} with no quantization. We fix the
decoding seed to $123$.\footnote{The random-direction specificity experiments
(Section~\ref{sec:analysis_random_direction}) for Gemma-2-9B, Llama-3.1-8B, Qwen2.5-1.5B,
and Qwen2.5-7B were run under an earlier configuration with seed $42$
(per-sample seed $42 + \text{sample\_id}$); all other generations use
seed $123$.}

Table~\ref{tab:models} lists every base model and its aligned counterpart
together with the exact HuggingFace identifiers used in our experiments.

\begin{table*}[t]
\centering
\small
\begin{tabular}{@{}lll@{}}
\toprule
\textbf{Family / stage} & \textbf{Base (HF id)} & \textbf{Aligned (HF id)} \\
\midrule
Llama-3.1       & \texttt{meta-llama/Meta-Llama-3.1-8B} & \texttt{meta-llama/Meta-Llama-3.1-8B-Instruct} \\
Qwen2.5-0.5B    & \texttt{Qwen/Qwen2.5-0.5B}  & \texttt{Qwen/Qwen2.5-0.5B-Instruct} \\
Qwen2.5-1.5B    & \texttt{Qwen/Qwen2.5-1.5B}  & \texttt{Qwen/Qwen2.5-1.5B-Instruct} \\
Qwen2.5-3B      & \texttt{Qwen/Qwen2.5-3B}    & \texttt{Qwen/Qwen2.5-3B-Instruct} \\
Qwen2.5-7B      & \texttt{Qwen/Qwen2.5-7B}    & \texttt{Qwen/Qwen2.5-7B-Instruct} \\
Qwen2.5-14B     & \texttt{Qwen/Qwen2.5-14B}   & \texttt{Qwen/Qwen2.5-14B-Instruct} \\
Qwen2.5-32B     & \texttt{Qwen/Qwen2.5-32B}   & \texttt{Qwen/Qwen2.5-32B-Instruct} \\
Gemma-2         & \texttt{google/gemma-2-9b}  & \texttt{google/gemma-2-9b-it} \\
Mistral         & \texttt{mistralai/Mistral-7B-v0.3} & \texttt{mistralai/Mistral-7B-Instruct-v0.3} \\
\midrule
OLMo-3 (SFT)    & \texttt{allenai/Olmo-3-7B}  & \texttt{allenai/Olmo-3-7B-Instruct-SFT} \\
OLMo-3 (DPO)    & \texttt{allenai/Olmo-3-7B}  & \texttt{allenai/Olmo-3-7B-Instruct-DPO} \\
OLMo-3 (Instruct) & \texttt{allenai/Olmo-3-7B} & \texttt{allenai/Olmo-3-7B-Instruct} \\
Tulu-3 (SFT)    & \texttt{meta-llama/Meta-Llama-3.1-8B} & \texttt{allenai/Llama-3.1-Tulu-3-8B-SFT} \\
Tulu-3 (DPO)    & \texttt{meta-llama/Meta-Llama-3.1-8B} & \texttt{allenai/Llama-3.1-Tulu-3-8B-DPO} \\
Tulu-3 (RLVR)   & \texttt{meta-llama/Meta-Llama-3.1-8B} & \texttt{allenai/Llama-3.1-Tulu-3-8B} \\
\bottomrule
\end{tabular}
\caption{Base and aligned model pairs with exact HuggingFace identifiers.
Tulu-3 and OLMo-3 are shown by post-training stage; the Tulu-3 RLVR
checkpoint is published under the bare repo name (no \texttt{-RLVR} suffix),
and the OLMo-3 final RLVR model is \texttt{Olmo-3-7B-Instruct}.}
\label{tab:models}
\end{table*}

\paragraph{Software Packages and Implementation Details.}
We implemented generation, activation extraction, and PASTA ablation in Python using PyTorch 2.4.1, HuggingFace Transformers 4.40--5.0, and Accelerate 0.30--1.0. Models were loaded with \texttt{AutoModelForCausalLM} and \texttt{AutoTokenizer}, using bfloat16 weights, no quantization, and \texttt{device\_map="auto"}. Generation used temperature $T=1.0$, top-$p=0.95$, \texttt{max\_new\_tokens=400}, and decoding seed 123; \texttt{top\_k}, \texttt{num\_beams}, and \texttt{repetition\_penalty} were left at their HuggingFace defaults. Prefix construction used \texttt{tiktoken} with the \texttt{cl100k\_base} encoding. PASTA direction extraction accumulates residual-stream means over generated-token positions and L2-normalizes the aligned-minus-base difference; ablation is implemented with custom PyTorch forward hooks at block-input, attention-output, and MLP-output sites.

\paragraph{Detector, Judge, and Statistics Implementations.}
For open-source detectors, we used Fast-DetectGPT with the \texttt{tiiuae/falcon-7b} and \texttt{tiiuae/falcon-7b-instruct} sampling/scoring pair, Binoculars with the same Falcon observer/performer models and \texttt{max\_token\_observed=512}, ImBD with the \texttt{xyzhu1225/ImBD-inference} LoRA adapter on \texttt{EleutherAI/gpt-neo-2.7B}, and RAIDAR with Llama-3.1-8B-Instruct rewriting followed by a scikit-learn MLP classifier over n-gram-overlap and fuzzy-ratio features. Commercial detector APIs were used for Pangram, GPTZero, and Originality.ai, with detector-specific probability scores thresholded at 0.5 unless otherwise specified in Appendix~D.1. Pairwise quality evaluation used Claude Sonnet 4.6 through AWS Bedrock with temperature 0.0 and \texttt{max\_tokens=8192}; prompts are provided in Appendix~F. Detector-quality confidence intervals use stratified percentile bootstrap with 2000 resamples, and pairwise judge-score confidence intervals use cluster bootstrap with 10000 resamples.

\subsection{Prompts}
\label{app:prompts}

This section describes how, for each of the 100 prefixes in each of
the 5 domains, we build a pair of prompts - one for the Base model
and one for the Aligned model - that elicit on-topic continuations
in each model's natural input format. The two prompts are derived from
the \emph{same} source record so that, after stripping the shared
opening from both outputs, the AI detector scores text generated under
matched topical conditions.

\paragraph{Design rationale: naturalistic prompting}
A central methodological choice is that we do \emph{not} feed Base and
Aligned models identical input. Pre-instruction-tuning Base models
expect raw text and degrade or copy boilerplate when given chat-style
instructions. Conversely, instruction-tuned Aligned models produce
assistant boilerplate (e.g.\ ``Sure! Here is\ldots'') when given a raw
text continuation. Identical prompts therefore handicap one model and
artifactually inflate or deflate detection rates.

We instead use what we call ``naturalistic'' prompting: the Base model receives a  document-style framing followed by the raw prefix, and the Aligned model receives a chat-formatted instruction plus an opening-anchor constraint derived from the same prefix. Both models are then decoded
with identical sampling parameters
($T = 1.0$, top-$p = 0.95$,
$\texttt{max\_new\_tokens} = 400$).

\paragraph{Common ingredients}

For every prefix record (Appendix~\ref{app:prefix-extraction}) the
template builder derives three fields:

\begin{itemize}
    \item \texttt{prefix} - the cl100k\_base-truncated source text
    (75 - 100 tokens, see Table~\ref{tab:domain-datasets}). Used
    verbatim by the Base model.
    \item \texttt{topic} / \texttt{instruction} / \texttt{clean\_prompt}
    - a short natural-language description of the writing task,
    derived from a metadata field of the source record (e.g.\
    CNN/DailyMail \texttt{highlights}, Medium \texttt{title}).
    Used by the Aligned model.
    \item \texttt{opening\_phrase} - the first 50 cl100k\_base tokens
    of \texttt{prefix}, snapped to the nearest word boundary. Used by the Aligned model as a
    \texttt{Begin with: "\ldots"} constraint, so that both model outputs
    start from the same opening context. After generation the opening
    is stripped from both Base and Aligned outputs prior to AI
    detection.
\end{itemize}

\paragraph{Per-domain templates}

Table~\ref{tab:prompt-templates} lists the exact templates for each
domain. \texttt{<prefix>} denotes the truncated prefix,
\texttt{<opening>} denotes the 50-token opening anchor, and the other
placeholders are the metadata-derived fields from
Section~\ref{app:prefix-extraction}.

\begin{table*}[h]
\centering
\caption{Naturalistic prompt templates per domain. The Base prompt is
sent as raw text; the Aligned prompt is delivered as a single
\texttt{user}-role chat message and rendered through the model's own
chat template at inference time.}
\label{tab:prompt-templates}
\small
\begin{tabular}{p{0.18\linewidth}p{0.36\linewidth}p{0.36\linewidth}}
\toprule
Domain & Base prompt & Aligned prompt (user message) \\
\midrule
College Essays &
\texttt{The following is an academic essay.\textbackslash{}n\textbackslash{}n<prefix>} &
\texttt{Write an academic essay about: <topic>\textbackslash{}n\textbackslash{}nBegin with: "<opening>"} \\
\addlinespace
News Articles &
\texttt{<prefix>} (CNN article body, self-framing) &
\texttt{Write a news article about the following: <highlights>\textbackslash{}n\textbackslash{}nBegin with: "<opening>"} \\
\addlinespace
Sci.\ Abstracts &
\texttt{Abstract:\textbackslash{}n<prefix>} &
\texttt{Write a scientific research abstract for a paper titled: <title>\textbackslash{}n\textbackslash{}nBegin with: "<opening>"} \\
\addlinespace
Creative Fiction &
\texttt{The following is a short story.\textbackslash{}n\textbackslash{}n<prefix>} &
\texttt{Write a short story based on this prompt: <clean\_prompt>\textbackslash{}n\textbackslash{}nBegin with: "<opening>"} \\
\addlinespace
Opinion Pieces &
\texttt{The following is an opinion piece.\textbackslash{}n\textbackslash{}n<prefix>} &
\texttt{Write an opinion piece about: <topic>\textbackslash{}n\textbackslash{}nBegin with: "<opening>"} \\
\bottomrule
\end{tabular}
\end{table*}

\paragraph{Notes on metadata-derived fields.}
\texttt{<topic>} for college essays is the URL slug from the source
IvyPanda permalink (de-hyphenated and title-cased; falls back to the
first line of the essay).
\texttt{<topic>} for opinion pieces is the Medium article title.
\texttt{<highlights>} for news is the journalist-written 1--3-bullet
summary in CNN/DailyMail.
\texttt{<title>} for scientific abstracts is the paper title from the
S2ORC record.
\texttt{<clean\_prompt>} for creative fiction is the original
\texttt{r/WritingPrompts} thread title with the leading bracketed tag
(\texttt{[WP]}, \texttt{[TT]}, \texttt{[EU]}, \ldots) stripped.

\paragraph{Worked example (news article \#0).}
\begin{itemize}
    \item \textbf{Base input:} the first $\sim$75 tokens of the article
    body, e.g.\ \textit{``It’s set to be one of the most groundbreaking
    moments in humanity’s six decades of space exploration: Today
    SpaceX was due to land a rocket on a floating platform after
    launching\ldots''}
    \item \textbf{Aligned input (chat user-turn):}
    \textit{``Write a news article about the following: SpaceX was
    planning to land a Falcon 9 rocket at Cape Canaveral today.
    The rocket was to attempt to land\ldots\textbackslash{}n\textbackslash{}n
    Begin with: ``It’s set to be one of the most groundbreaking moments
    in humanity’s six decades of space exploration\ldots''}
\end{itemize}

\section{Additional Details on H1}
\label{app:h1_details}

\subsection{Additional Corpus-Affinity Results}
\label{sec:posttraining_appendix}

This appendix reports the per-domain scores underlying H1 in
Section~\ref{sec:posttraining}. Across five source-text domains and
two corpus-affinity metrics, the mean ordering Human $>$ Base $>$ Aligned holds
in every domain. Tables~\ref{tab:appendix-density} and
\ref{tab:appendix-coverage} report both raw scores and scores normalized by the
within-domain human mean, so Human $=1.000$ by construction.

\paragraph{Metric Definitions}
\label{subsec:posttraining_metrics}

For each continuation, we tokenize into words and compute a per-position
longest-match vector $L$, where $L[i]$ is the length, in words, of the longest
verbatim span starting at position $i$ that appears in the RedPajama+ index.

\paragraph{Overlap Density.}
Density averages longest-match lengths over all word positions:
\[
\text{Density} \;=\; \frac{1}{N} \sum_{i=1}^{N} L[i],
\]
where $N$ is the number of words in the continuation. Higher density means that
more positions begin longer corpus-matched spans. We use density as the primary
corpus-affinity metric because it aggregates evidence across all positions
rather than relying on a single maximum match.

\paragraph{Coverage@10.}
Coverage@10 is the fraction of words covered by verbatim corpus-matched spans
of length at least 10. We compute it greedily from left to right: if
$L[i]\geq 10$, the next $L[i]$ words are marked covered and the scan advances by
$L[i]$; otherwise it advances by one word. Unlike density, Coverage@10 ignores
short matches and focuses only on long, contiguous verbatim overlap.

Both density and Coverage@10 follow the same mean ordering across all five
domains: Human $>$ Base $>$ Aligned. The gap is largest when the RedPajama+
index has close source overlap with the evaluation domain, such as scientific
abstracts, creative fiction, college essays, and news articles. Opinion pieces
show a weaker density gap and weaker per-example ordering, consistent with less
direct source coverage. Thus, the direction of the effect is consistent across
domains, while its statistical strength varies with source coverage and metric
choice.

\begin{table*}[t]
\centering
\small
\caption{
\textbf{Post-training reduces verbatim corpus affinity.}
For matched Human/Base/Aligned continuations from the same prefix, infini-gram density measures the mean longest RedPajama+-matched span across word positions. Across domains, density consistently follows Human $>$ Base $>$ Aligned. Normalized scores divide by the within-domain human mean to show the relative magnitudes.
}
\label{tab:appendix-density}
\begin{tabular}{l ccc ccc}
\toprule
                                  & \multicolumn{3}{c}{\textbf{Overlap Density}} & \multicolumn{3}{c}{\textbf{Normalized Overlap Density}} \\
\cmidrule(lr){2-4} \cmidrule(lr){5-7}
\textbf{Domain}                            & \textbf{Human}      & \textbf{Base}      & \textbf{Aligned}      & \textbf{Human}      & \textbf{Base}      & \textbf{Aligned}      \\
\midrule
Scientific abstracts              & 12.22  & 8.82   & 6.25   & 1.000  & 0.722  & 0.511  \\
Creative fiction                  & 10.27  & 6.76   & 5.16   & 1.000  & 0.658  & 0.502  \\
Opinion pieces                    & 6.11   & 5.11   & 4.78   & 1.000  & 0.836  & 0.782  \\
College essays                    & 10.73  & 5.70   & 4.87   & 1.000  & 0.531  & 0.454  \\
News articles                     & 11.74  & 7.78   & 5.34   & 1.000  & 0.663  & 0.455  \\
\bottomrule
\end{tabular}
\end{table*}

\begin{table*}[t]
\centering
\small
\setlength{\tabcolsep}{5pt}
\renewcommand{\arraystretch}{1.12}
\caption{
\textbf{Post-training reduces long-span verbatim overlap.}
For matched Human/Base/Aligned continuations from the same prefix, Coverage@10 measures the fraction of words covered by RedPajama+-matched spans of at least 10 words. Across domains, coverage follows Human $>$ Base $>$ Aligned. Normalized scores divide by the within-domain human mean.
}
\label{tab:appendix-coverage}
\begin{tabular}{@{}l ccc ccc@{}}
\toprule
& \multicolumn{3}{c}{\textbf{Coverage@10}} 
& \multicolumn{3}{c}{\textbf{Normalized Coverage@10}} \\
\cmidrule(lr){2-4} \cmidrule(lr){5-7}
\textbf{Domain} 
& \textbf{Human} & \textbf{Base} & \textbf{Aligned} 
& \textbf{Human} & \textbf{Base} & \textbf{Aligned} \\
\midrule
Scientific Abstracts & 0.647 & 0.375 & 0.237 & 1.000 & 0.580 & 0.366 \\
Creative Fiction     & 0.676 & 0.264 & 0.118 & 1.000 & 0.391 & 0.175 \\
Opinion Pieces       & 0.185 & 0.080 & 0.045 & 1.000 & 0.432 & 0.243 \\
College Essays       & 0.517 & 0.126 & 0.080 & 1.000 & 0.244 & 0.155 \\
News Articles        & 0.550 & 0.250 & 0.120 & 1.000 & 0.455 & 0.218 \\
\bottomrule
\end{tabular}
\end{table*}

\subsection{Pretraining-style reference index}
\label{app:ngram-index}

This appendix details the infini-gram reference corpus $C$ used in Section~\ref{sec:posttraining} to measure verbatim source affinity. The index is intended as a proxy for presence in pretraining data. All sources are restricted to content authored on or before December 2022 wherever source-side filtering is feasible. The final index covers approximately 3.5 TB of raw text, corresponding to roughly 600B LLaMA-2-tokenized tokens, and is split into 16 shards.

\subsection{Indexed sources}
\label{app:ngram-index:sources}

The corpus contains 14 sources, grouped by their primary contributing domain:

\begin{itemize}[leftmargin=*]
    \item \textbf{Scientific text}: \texttt{allenai/peS2o} V1 train (papers $\leq$ 2022-12-01), RedPajama-Data-1T arXiv, and PubMed Central Open Access articles filtered to earliest publication year $\leq 2022$. The peS2o source is included as a verbatim-match upper bound because it is the dataset version underlying the scientific-abstract source corpus.
    \item \textbf{Reddit and discussion}: Pushshift Reddit comments and submissions from the Internet Archive dump \texttt{pushshift\_reddit\_200506\_to\_202212}, covering RC 2005-12 to RC 2022-12 and RS 2005-06 to RS 2022-12.
    \item \textbf{Narrative fiction}: BookCorpusOpen (\texttt{lucadiliello/bookcorpusopen}), corresponding to the 2015 self-published fiction reissue.
    \item \textbf{News}: \texttt{vblagoje/cc\_news}, covering CC-News from January 2017 to December 2019, and sampled CC-News WARC files from Common Crawl for January 2020 to December 2022.
    \item \textbf{College essays}: IvyPanda essays recovered via the Wayback Machine CDX API, yielding 55,434 unique URLs collapsed by \texttt{urlkey} from 2018 to 2022-12-31.
    \item \textbf{General web}: RedPajama-Data-1T Common Crawl, AllenAI C4 English, and RedPajama-Data-1T Wikipedia, StackExchange, and GitHub.
\end{itemize}

Table~\ref{tab:indexed-datasets} reports the per-source processed sizes.

\begin{table*}[t]
\centering
\small
\setlength{\tabcolsep}{4pt}
\caption{Datasets included in the retrieval index used for pre-training overlap analysis. Sizes are approximate and correspond to the processed subsets stored locally.}
\label{tab:indexed-datasets}
\begin{tabular}{p{3.5cm} p{9.0cm} p{1.0cm}}
\toprule
\textbf{Dataset} & \textbf{Description} & \textbf{Size} \\
\midrule

\texttt{arxiv} &
Scientific papers from the RedPajama-1T ArXiv subset. &
88 GB \\

\texttt{c4} &
English split of the Colossal Clean Crawled Corpus (C4). &
305 GB \\

\texttt{common\_crawl} &
RedPajama Common Crawl slice spanning snapshots from 2019--2023. &
1.7 TB \\

\texttt{github} &
Source code from public GitHub repositories. &
213 GB \\

\texttt{stackexchange} &
Question-answer content from StackExchange forums. &
75 GB \\

\texttt{wikipedia} &
Wikipedia articles from RedPajama-1T. &
112 GB \\

\texttt{peS2o} &
Scientific and technical documents from the peS2o corpus. &
94 GB \\

\texttt{bookcorpusopen} &
Open version of BookCorpus containing fiction books. &
2.2 GB \\

\texttt{cc\_news} &
CC-News articles from 2017--2019. &
0.5 GB \\

\texttt{cc\_news\_2020\_2022} &
Additional CC-News articles from 2020--2022. &
33 GB \\

\texttt{ivypanda} &
Educational essays from IvyPanda. &
166 MB \\

\texttt{pmc} &
PubMed Central Open Access biomedical articles. &
30 GB \\

\texttt{pushshift\_comments} &
Reddit comments from 2005--2022. &
168 TB \\

\texttt{pushshift\_submissions} &
Reddit submissions from 2005--2022. &
191 TB \\

\bottomrule
\end{tabular}
\end{table*}

\subsection{Index implementation}
\label{app:ngram-index:impl}

We use the open-source \texttt{infini\_gram} Python package, which builds a per-token suffix array over a LLaMA-2-tokenized byte stream. We tokenize with the LLaMA-2-7B-HF fast tokenizer using \texttt{use\_fast=True}. A one-time validation against the slower SentencePiece tokenizer on 50 random RedPajama-arXiv records confirmed byte-identical token IDs. The fast path is approximately 2.5 times faster per worker.

\subsection{Source-side preprocessing}
\label{app:ngram-index:preprocessing}

When a source is not already line-delimited JSON with a text field, we convert it at ingest. Pushshift dumps are streamed with \texttt{zstandard.ZstdDecompressor}; comments emit \texttt{body}, while submissions emit \texttt{title + "\textbackslash n\textbackslash n" + selftext}. Rows with deleted or removed text, or fewer than one character of text, are dropped. PMC XML tarballs are streamed with \texttt{tarfile}; for each article, we concatenate \texttt{article-title}, \texttt{abstract}, and \texttt{body}, joined by blank lines, and drop articles with earliest publication year after 2022 or fewer than 200 characters of body text. CC-News WARC files and IvyPanda Wayback snapshots are converted from HTML using \texttt{trafilatura} with comments and tables disabled; extracted text shorter than 200 characters is discarded. HuggingFace parquet sources such as BookCorpusOpen and CC-News are re-encoded as \texttt{jsonl.zst} with one text field per row.

\subsection{Query-time scoring}
\label{app:ngram-index:scoring}

Each evaluation row is scored against the index by computing, for every word position $i$, the longest contiguous span $\ell_i \in \mathbb{N}$ such that the substring $w_i,w_{i+1},\ldots,w_{i+\ell_i-1}$ has count at least one in the index. Single-character span lookup uses the \texttt{count} query of the local infini-gram server; longer spans are found with binary search, bracketed by the 1000-character LLaMA-tokenizer query cap.

\subsection{Surface-form normalization}
\label{app:ngram-index:normalization}

Exact substring lookup is sensitive to superficial formatting differences. For example, some test rows store quotation marks in a Penn-Treebank-style tokenized form, while Pushshift uses straight ASCII quotes and peS2o, PMC, and IvyPanda preserve curly typography. To reduce false mismatches, we compute $\ell_i$ under up to three surface forms of each row:
\begin{enumerate}[leftmargin=*]
    \item \texttt{orig}: the test row as-is;
    \item \texttt{ptb\_only}: NFC normalization, with Penn-Treebank quote tokens replaced by standard quote characters while preserving curly typographic quotes;
    \item \texttt{full\_ascii}: NFC normalization plus conversion of Penn-Treebank quotes, curly quotes, en-dashes, em-dashes, ellipses, and non-breaking spaces into ASCII equivalents.
\end{enumerate}
All normalizers preserve word boundaries, so positions align across forms. For each position, we report:
\[
\ell_i = \max_f \ell_i^{(f)}.
\]

For the count of each reported span in the corpus, we sum the per-form counts after deduplicating query strings:
\[
\textrm{count}(\textrm{span}_i) = \sum_{q \in \{w_i^{(f)} \cdots w_{i+\ell_i-1}^{(f)}\}_f} \textrm{count}_{\textrm{idx}}(q),
\]
where the set is constructed from distinct query strings only. This captures cases where curly and ASCII forms match disjoint subsets of the corpus, while avoiding double-counting when all forms collapse to the same string.

For each row, we report \texttt{coverage@10}, \texttt{max\_match\_words}, \texttt{max\_match\_normalized}, and \texttt{inf\_gram\_density\_words}, together with all coverage spans of length at least 10 and their per-corpus counts.

\section{Additional Details on H2}
\label{app:h2_details}

\subsection{Detector Details}
\label{app:detector-details}

We evaluate seven AI-text detectors spanning closed-source commercial
systems and open-source methods. The three commercial detectors are
Pangram~\citep{emi2024technical}, GPTZero,\footnote{\url{https://gptzero.me}}
and Originality.ai.\footnote{\url{https://originality.ai}} The four
open-source detectors are ImBD~\citep{chen2025imitate}, Fast-DetectGPT~\citep{bao2024fastdetect},
Binoculars~\citep{hans2024binoculars}, and RAIDAR~\citep{mao2024raidar}.
We treat all detector outputs as signals of stylistic separability rather
than ground-truth authorship labels.

For binary flagging rates we use each detector's canonical decision
threshold: Binoculars flags a text as machine-generated when its score is
below $0.85$, and ImBD flags when $\mathrm{prob\_machine} \geq 0.5$.
RAIDAR has no upstream threshold and is reported as the mean
\texttt{raidar\_score} in raw points (pts). For Pangram, GPTZero, and
Originality.ai we use the providers' reported AI-probability outputs at the
$\mathrm{fraction\_ai} > 0.5$ threshold.

\subsection{Evaluating AI-detector quality}
\label{sec:ai_detector_quality_eval}

We evaluate detector quality using a fixed-threshold aligned-vs.-human classification accuracy over the same five domains and eight model families used in H2. Table~\ref{tab:macro-accuracy-main-pipeline-final} reports per-domain averages, per-model averages, and the grand macro-average across (domain, model) combinations. Ranking by the grand score, Pangram is the strongest detector ($\bar{x}=0.982$), followed by GPTZero ($0.909$); ImBD and Originality.ai are tied next ($0.897$), followed by other open-source detectors. We found that although Originality.ai has a higher true AI detection rate, it also has a higher false AI detection rate for human-generated text, indicating a bias toward classifying text as AI-generated.
We use fixed-threshold accuracy over ranking-based metric such as AUROC in order to evaluate absolute AI-detection rate of human and aligned model generations rather than relative ranking between human and aligned generations done by AUROC.
As a robustness check, we repeated the detector-quality evaluation on a length-controlled subset and found that the detector ranking did not change.

\begin{table*}[t]
\centering
\scriptsize
\setlength{\tabcolsep}{3pt}
\caption{\textbf{Final combined macro-accuracy table: per-domain marginals ($N=5$), per-model marginals ($M=8$), and the grand mean.} Grand ($\bar{x}$) is the macro-averaged accuracy across all domain-and-model combinations. Pangram and GPTZero are good detectors based on grand accuracy. CE = College essays, CF = Creative fiction, NA = News articles, OP = Opinion pieces, SA = Scientific abstracts. Q3 / Q7 / Q14 = Qwen-2.5-\{3,7,14\}B-Instruct.}
\label{tab:macro-accuracy-main-pipeline-final}
\begin{tabular}{l ccccc cccccccc c}
\toprule
 & \multicolumn{5}{c}{Per domain ($\bar{x}_{\text{model}}(i)$)} & \multicolumn{8}{c}{Per model ($\bar{x}_{\text{domain}}(j)$)} & \multicolumn{1}{c}{Grand} \\
\cmidrule(lr){2-6} \cmidrule(lr){7-14} \cmidrule(lr){15-15}
Detector & CE & CF & NA & OP & SA & Llama & Tulu & Gemma & Mistral & OLMo & Q3 & Q7 & Q14 & $\bar{x}$ \\
\midrule
Pangram        & 1.000 & 1.000 & 0.983 & 0.992 & 0.600 & 0.920 & 1.000 & 1.000 & 1.000 & 0.988 & 0.979 & 0.984 & 1.000 & 0.982 \\
GPTZero        & 0.994 & 0.870 & 0.927 & 0.885 & 0.600 & 0.887 & 0.902 & 0.969 & 0.970 & 0.916 & 0.865 & 0.935 & 0.837 & 0.909 \\
IMBD           & 0.800 & 0.984 & 0.988 & 0.854 & 0.600 & 0.847 & 0.903 & 0.919 & 0.919 & 0.863 & 0.903 & 0.919 & 0.919 & 0.897 \\
Originality    & 0.950 & 0.716 & 0.994 & 0.938 & 0.800 & 0.894 & 0.916 & 0.909 & 0.916 & 0.834 & 0.909 & 0.883 & 0.909 & 0.897 \\
fast\_detectgpt & 0.900 & 0.895 & 0.826 & 0.938 & 0.400 & 0.807 & 0.869 & 0.892 & 0.932 & 0.856 & 0.886 & 0.917 & 0.857 & 0.875 \\
Binoculars     & 0.938 & 0.846 & 0.657 & 0.859 & 0.500 & 0.870 & 0.815 & 0.964 & 0.975 & 0.594 & 0.681 & 0.884 & 0.725 & 0.815 \\
RAIDAR$^{\dagger}$ & 0.500 & 0.613 & 0.500 & 0.554 & 0.542 & 0.542 & 0.543 & 0.547 & 0.543 & 0.547 & 0.531 & 0.533 & 0.547 & 0.542 \\
\bottomrule
\end{tabular}

\vspace{2pt}
\end{table*}

\subsection{H2 Results Breakdown by Domain}
\label{app:h2-breakdown-domain}

\begin{figure*}[t]
\centering
\includegraphics[width=\textwidth]{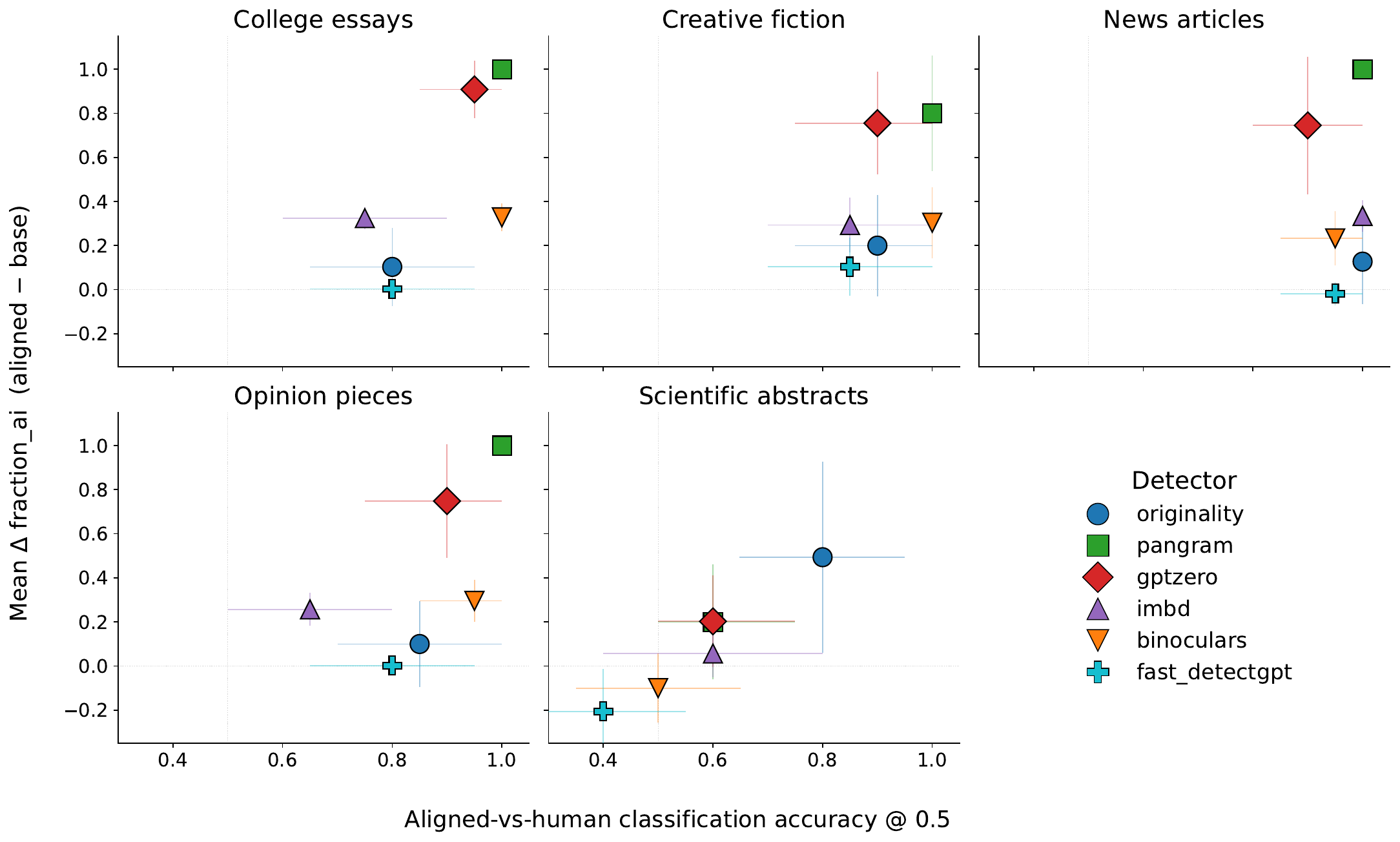}
\caption{$\Delta$(aligned-base) vs. detector accuracy per data domain. Each point is one detector identified by a color and marker. The horizontal/vertical bars show 95\% confidence intervals. \Pangram and \GPTZero consistently occupy top-right corner: they classify human vs. aligned cleanly and produce a large post-training $\Delta$. \RAIDAR consistently leads to low accuracy classification. Originality and other OSS detectors' accuracy fall in between.}
\label{fig:accuracy-vs-delta-by-domain}
\end{figure*}

Figure~\ref{fig:accuracy-vs-delta-by-domain} breaks down the relationship between detector quality and the aligned--base detection gap by domain. Across domains, Pangram and GPTZero generally lie in the high-accuracy, high-$\Delta$ region, indicating that the detectors with the best aligned-vs.-human accuracy are also the ones that most clearly expose the post-training shift from base to aligned generations. Lower-accuracy detectors show weaker or less stable aligned--base gaps, with RAIDAR consistently near the low-accuracy region and Originality.ai showing more domain variation, especially in creative fiction. These domain-level patterns support the use of Pangram and GPTZero as the primary high-accuracy detectors in the main H2 analysis.

\subsection{H2 Results Breakdown by Models}
\label{app:h2-breakdown-models}

\begin{figure*}[t]
\centering
\includegraphics[width=\textwidth]{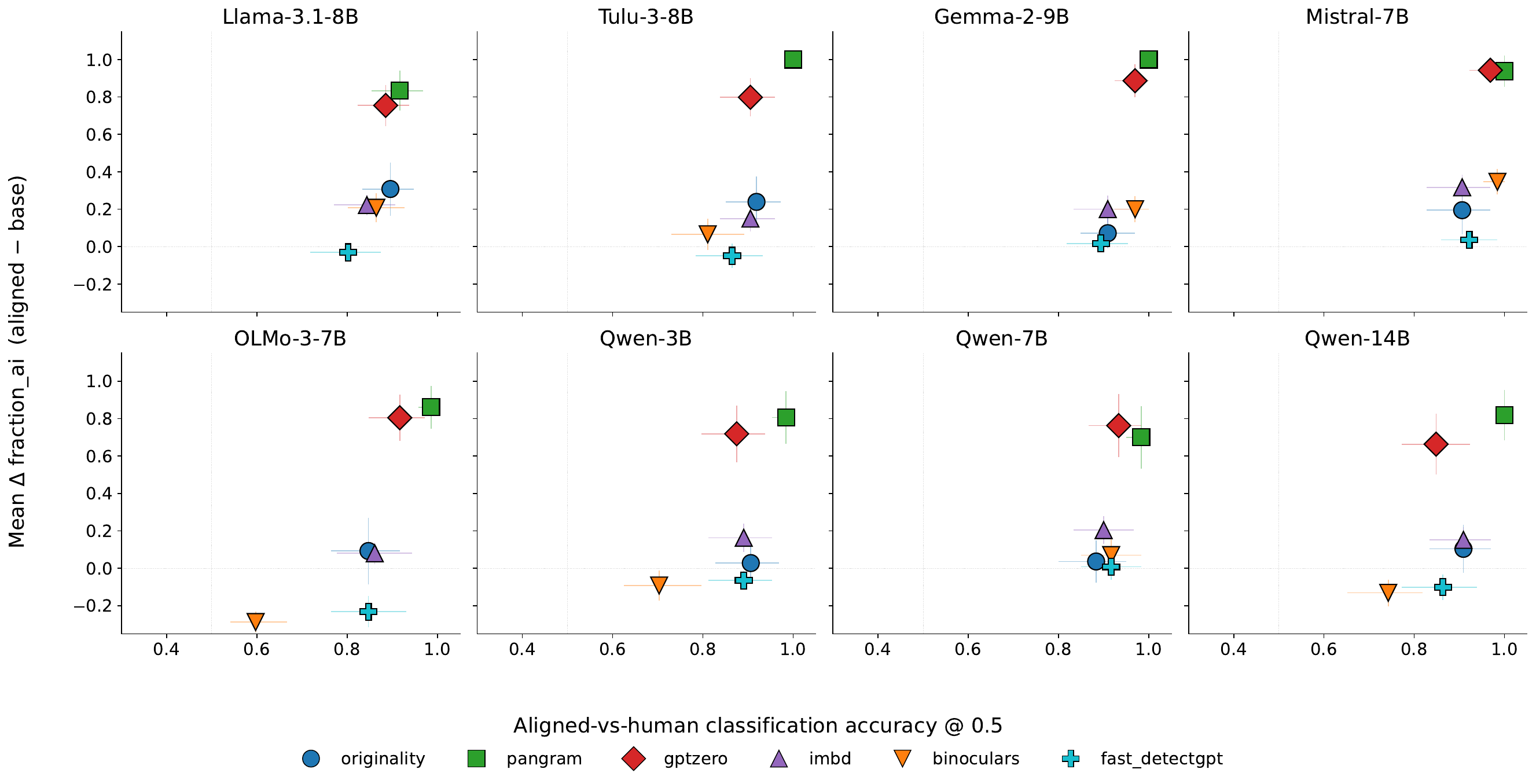}
\caption{$\Delta$(aligned-base) vs. detector quality (balanced accuracy) per base-aligned model pair.}
\label{fig:delta-vs-quality-by-model-accuracy}
\end{figure*}

Figure~\ref{fig:delta-vs-quality-by-model-accuracy} shows the same relationship between detector quality and the aligned--base detection gap, now broken down by base--aligned model pair. The model-level view shows that H2 is not driven by a single model family: Pangram and GPTZero remain among the most accurate detectors and usually produce the largest positive aligned--base gaps across models. In contrast, lower-ranked detectors from Table~\ref{tab:macro-accuracy-main-pipeline-final} yield smaller or less consistent gaps, suggesting that apparent failures of H2 for some detectors are better interpreted as detector-quality limitations rather than absence of a post-training shift.

\section{\pasta Details}
\label{app:pasta_details}

\subsection{Algorithm}
\label{app:pasta-algo}

This appendix provides pseudocode for the three components of \pasta. Algorithm~\ref{alg:pasta-direction} extracts a candidate AI-detection direction $v_{\text{cross}}[\ell]$ at every layer $\ell$ via a difference-in-means between the Aligned and Base model on naturalistic prompts. Algorithm~\ref{alg:pasta-ablate} performs inference-time directional ablation by projecting a fixed unit direction $\hat v$ out of every residual-stream write site during autoregressive decoding. Algorithm~\ref{alg:pasta-sweep} composes the previous two: it sweeps the candidate layer $\ell$, generates ablated continuations, scores them with a downstream detector, and returns the causally most effective layer $\ell^*$.

\begin{algorithm}[t]
\caption{Cross-Model Direction Extraction}
\label{alg:pasta-direction}
\begin{algorithmic}[1]
\Require Aligned model $\mathcal{M}_{\text{a}}$; Base model $\mathcal{M}_{\text{b}}$; aligned-format prompts $\mathcal{P}_{\text{a}}$; base-format prompts $\mathcal{P}_{\text{b}}$ ($|\mathcal{P}_{\text{a}}| = |\mathcal{P}_{\text{b}}| = N$); generation hyperparameters $T$; layer count $L_{\text{tot}}$; residual width $d$.
\Ensure Unit-norm directions $\{v_{\text{cross}}[\ell]\}_{\ell=0}^{L_{\text{tot}}-1}$.
\For{$(\mathcal{M}, \mathcal{P}) \in \{(\mathcal{M}_{\text{a}}, \mathcal{P}_{\text{a}}),\, (\mathcal{M}_{\text{b}}, \mathcal{P}_{\text{b}})\}$}
  \State $s[\ell] \gets \mathbf{0} \in \mathbb{R}^{d}$ \textbf{(fp64)} for $\ell = 0,\ldots,L_{\text{tot}}-1$; $\;n \gets 0$
  \For{each prompt $p \in \mathcal{P}$}
    \State $g_p \gets \textsc{Generate}(\mathcal{M}, p, T)$ \Comment{natural continuation}
    \State $\mathrm{ids} \gets \text{concat}(\textsc{Tok}(p),\, \textsc{Tok}(g_p))$
    \State $H \gets \textsc{Forward}(\mathcal{M}, \mathrm{ids};\,\text{hidden\_states})$ \Comment{$H[\ell] \in \mathbb{R}^{|\mathrm{ids}|\times d}$}
    \State $\mathrm{st} \gets |\textsc{Tok}(p)| - 1$;\quad $\mathrm{en} \gets \mathrm{st} + |\textsc{Tok}(g_p)|$
    \For{$\ell = 0,\ldots,L_{\text{tot}}-1$}
      \State $s[\ell] \gets s[\ell] + \sum_{i=\mathrm{st}}^{\mathrm{en}-1} H[\ell][i]$
    \EndFor
    \State $n \gets n + |\textsc{Tok}(g_p)|$
  \EndFor
  \State $\mu_{\mathcal{M}}[\ell] \gets s[\ell] / n$ for $\ell = 0,\ldots,L_{\text{tot}}-1$ \Comment{per-model mean}
\EndFor
\For{$\ell = 0,\ldots,L_{\text{tot}}-1$}
  \State $\Delta[\ell] \gets \mu_{\mathcal{M}_{\text{a}}}[\ell] - \mu_{\mathcal{M}_{\text{b}}}[\ell]$
  \State $v_{\text{cross}}[\ell] \gets \Delta[\ell] / \lVert \Delta[\ell] \rVert_2$
\EndFor
\State \Return $\{v_{\text{cross}}[\ell]\}_{\ell=0}^{L_{\text{tot}}-1}$
\end{algorithmic}
\end{algorithm}

\begin{algorithm}[t]
\caption{Inference-Time Directional Ablation}
\label{alg:pasta-ablate}
\begin{algorithmic}[1]
\Require Aligned model $\mathcal{M}_{\text{a}}$ with $L_{\text{tot}}$ transformer blocks; unit direction $v \in \mathbb{R}^{d}$ with $\|v\|_2=1$; ablation strength $\alpha \ge 0$ with default $\alpha=1$; prompt $p$; generation hyperparameters $T$.
\Ensure Detection-ablated generation $g$.
\For{$\ell = 0,\ldots,L_{\text{tot}}-1$} \Comment{register $3 L_{\text{tot}}$ hooks}
  \State \textbf{pre-hook} on block $\ell$:\; $x \gets x - \alpha (x \cdot v)\,v$
  \State \textbf{post-hook} on attention output of block $\ell$:\; $x \gets x - \alpha (x \cdot v)\,v$
  \State \textbf{post-hook} on MLP output of block $\ell$:\; $x \gets x - \alpha (x \cdot v)\,v$
\EndFor
\State $g \gets \textsc{Generate}(\mathcal{M}_{\text{a}}, p, T)$ \Comment{hooks active throughout decoding}
\State Remove all hooks from $\mathcal{M}_{\text{a}}$
\State \Return $g$
\end{algorithmic}
\end{algorithm}

\begin{algorithm}[t]
\caption{Layer Sweep for Identifying $\ell^*$}
\label{alg:pasta-sweep}
\begin{algorithmic}[1]
\Require Aligned model $\mathcal{M}_{\text{a}}$; directions $\{v_{\text{cross}}[\ell]\}_{\ell=0}^{L_{\text{tot}}-1}$ from Alg.~\ref{alg:pasta-direction}; evaluation prompts $\mathcal{P}_{\text{eval}}$; generation hyperparameters $T$; AI-text detector $\mathcal{D}$ (e.g., Pangram).
\Ensure Causally most effective layer $\ell^*$ and sweep $\{d[\ell]\}_{\ell=0}^{L_{\text{tot}}-1}$.
\For{$\ell = 0,\ldots,L_{\text{tot}}-1$}
  \State $\mathcal{G} \gets \emptyset$
  \For{each prompt $p \in \mathcal{P}_{\text{eval}}$}
    \State $g \gets \textsc{AblatedGenerate}(\mathcal{M}_{\text{a}},\, v_{\text{cross}}[\ell],\, p,\, T)$ \Comment{Alg.~\ref{alg:pasta-ablate}}
    \State $\mathcal{G} \gets \mathcal{G} \cup \{g\}$
  \EndFor
  \State $d[\ell] \gets \mathcal{D}(\mathcal{G})$ \Comment{mean fraction flagged AI}
\EndFor
\State $\ell^* \gets \arg\min_{\ell \in \{0,\ldots,L_{\text{tot}}-1\}} d[\ell]$
\State \Return $\ell^*,\, \{d[\ell]\}_{\ell=0}^{L_{\text{tot}}-1}$
\end{algorithmic}
\end{algorithm}

\section{Additional Details on Quality of \pasta}
\label{app:qual_details}

\subsection{Prompts for LLM-Judge}
\label{app:judge_prompts}

For completeness, we report verbatim the two LLM-as-judge prompts used to score \pasta and baseline generations on writing fluency and prompt relevance. All the judgments were collected with \ClaudeSonnet{} at temperature $0$ via the Anthropic API. To control for any first-position bias of the judge, every (TEXT~A, TEXT~B) pair is evaluated in both the AB and BA orderings, and the resulting per-text scores are pooled before aggregation.

Figures~\ref{fig:judge-prompt-fluency-a}, \ref{fig:judge-prompt-fluency-b}, and~\ref{fig:judge-prompt-fluency-c} show our fluency judge prompt. It checks whether each response is easy to read as text: grammatical, coherent, natural in phrasing, and free of disruptions that impede comprehension, without judging whether the response is relevant, factual, or stylistically preferable.
Figures~\ref{fig:judge-prompt-relevance-a}, \ref{fig:judge-prompt-relevance-b}, and~\ref{fig:judge-prompt-relevance-c} show our prompt-relevance judge prompt.
The relevance prompt checks whether each response satisfies the user's task: it evaluates alignment with the requested topic, genre or form, and explicit instructions, without judging writing polish or fluency.

\begin{figure*}[!ht]
\centering
\fbox{%
\begin{minipage}{0.95\textwidth}
\small\ttfamily\raggedright
\setlength{\parindent}{0pt}%
\setlength{\parskip}{0.4em}%
\textbf{\normalsize\rmfamily\upshape Fluency judge prompt (part 1 of 3)}\par\smallskip\hrule\smallskip
You are comparing two model responses only on FLUENCY.

Fluency means whether each response is grammatical, coherent, and natural to read. Judge only the form of the writing --- how it reads --- not its content, topic, accuracy, relevance, genre appropriateness, persuasiveness, creativity, or factuality.

Your task is to decide which response is more fluent.

Rate each response on this 1--5 fluency scale:

1 --- Severely non-fluent. The response is so broken, garbled, or incoherent that it is very hard to read at all.

2 --- Mostly non-fluent. The response has frequent grammar, syntax, wording, or coherence problems that make reading effortful throughout.

3 --- Somewhat fluent. The response is understandable, but noticeable problems make the reader slow down, re-read, or lose the thread in places.

4 --- Fluent. The response may contain minor awkwardness or small errors, but nothing substantially slows reading or impedes comprehension.

5 --- Fully fluent. The response reads naturally from start to finish; grammar, syntax, phrasing, and flow do not make the reader pause.

What should make a response more fluent:

- It is grammatical and easy to read.

- Sentences are well formed.

- The wording is natural rather than awkward.

- The flow between sentences is coherent.

- The reader can follow the text without slowing down or re-reading.

- The text avoids garbled, corrupted, or non-prose stretches.

What should make a response less fluent:

- Grammatical errors, broken syntax, run-on syntax, missing words, or agreement errors.

- Awkward or unnatural phrasing.

- Choppy, disjointed, or hard-to-follow flow.

- Garbled text, corrupted text, stray quotation marks or backticks, visible escape characters such as "\textbackslash{}n" or "\textbackslash{}"", or dumped prompt/instruction fragments.

- Repetition or looping that disrupts readability.

- Abrupt incoherence that makes the reader lose the thread.
\end{minipage}%
}
\caption{LLM-as-judge prompt for pairwise \emph{fluency} scoring (part 1 of 3; continued in Figures~\ref{fig:judge-prompt-fluency-b} and~\ref{fig:judge-prompt-fluency-c}). Part 1 specifies the task framing (independent absolute scoring of two responses, with a separate pairwise winner derived from the scores), the 1--5 rubric anchored on how much problems impede reading, and the lists of attributes that make a response more or less fluent.}
\label{fig:judge-prompt-fluency-a}
\end{figure*}

\begin{figure*}[!ht]
\centering
\fbox{%
\begin{minipage}{0.95\textwidth}
\small\ttfamily\raggedright
\setlength{\parindent}{0pt}%
\setlength{\parskip}{0.4em}%
\textbf{\normalsize\rmfamily\upshape Fluency judge prompt (part 2 of 3)}\par\smallskip\hrule\smallskip
Focus only on fluency. Do NOT judge:

- Prompt relevance.

- Whether the response answers the user's request.

- Factual accuracy.

- Domain correctness.

- Persuasiveness.

- Creativity.

- Helpfulness.

- Informativeness.

- Depth.

- Specificity.

- Verbosity.

- Whether the response is too long or too short.

- Whether the response sounds formal, academic, polished, casual, conversational, or human-written.

Important clarification about relevance: Do not penalize a response for being off-topic, incomplete, or failing the prompt unless that failure creates a fluency problem in the visible text itself. Relevance is evaluated separately.

Important clarification about factuality: Do not penalize false, implausible, or unsupported claims when judging fluency. A factually wrong sentence can still be fluent if it reads smoothly.

Important clarification about register and style: Do not penalize informal, colloquial, plain, simple, or stylistically unusual prose if it is grammatical and easy to read. "Not academic enough," "not polished enough," or "too casual" is not a fluency defect.

Important clarification about technical density: Dense, jargon-heavy, or specialized prose, such as a scientific abstract, can be fully fluent if it is grammatical and coherent. Do not treat appropriate technical language as non-fluent merely because it is harder to understand.

Important clarification about formatting: Intentional, well-formed headings, lists, markdown, or paragraph breaks are formatting choices, not fluency defects. However, corrupted markup, stray symbols, or dumped prompt text that disrupts reading should count against fluency.

Important clarification about truncation: Judge only the visible content. Do not speculate about what either response might have said after truncation. Do not penalize an abrupt ending by itself. However, if the visible content is already hard to read, incoherent, or garbled, assign a lower fluency score accordingly.

Important clarification about minor issues: A single awkward phrase, or a few scattered minor slips that you can read straight past, should usually receive a 4, not a 3. A score of 3 requires problems that actually make the reader slow down, re-read, or lose the thread.

Important clarification about ties: Use "tie" when both responses are similarly fluent or similarly non-fluent. Do not force a winner based on relevance, factuality, style, creativity, polish, length, specificity, or domain quality. If both responses are fully fluent, choose "tie". If both responses have comparable fluency problems, choose "tie".
\end{minipage}%
}
\caption{LLM-as-judge prompt for pairwise \emph{fluency} scoring (part 2 of 3; continued in Figure~\ref{fig:judge-prompt-fluency-c}). Part 2 specifies the exclusions (what the judge must \emph{not} consider, including prompt relevance, factual accuracy, helpfulness, length, and register/human-likeness) and the eight clarifications (relevance, factuality, register and style, technical density, formatting, truncation, minor issues, ties) that disambiguate edge cases.}
\label{fig:judge-prompt-fluency-b}
\end{figure*}

\begin{figure*}[!ht]
\centering
\fbox{%
\begin{minipage}{0.95\textwidth}
\small\ttfamily\raggedright
\setlength{\parindent}{0pt}%
\setlength{\parskip}{0.4em}%
\textbf{\normalsize\rmfamily\upshape Fluency judge prompt (part 3 of 3)}\par\smallskip\hrule\smallskip
First assess Response A's fluency. Then assess Response B's fluency. Then compare them. Only after those assessments, assign scores and choose a winner.

Score/winner consistency rules:

- If Response A is meaningfully more fluent, assign Response A a higher fluency score and set winner to "A".

- If Response B is meaningfully more fluent, assign Response B a higher fluency score and set winner to "B".

- If both responses are similarly fluent, assign the same fluency score and set winner to "tie".

- If response\_a\_score > response\_b\_score, winner must be "A".

- If response\_b\_score > response\_a\_score, winner must be "B".

- If response\_a\_score = response\_b\_score, winner must be "tie".

- Do not assign different scores for tiny differences that do not meaningfully affect fluency.

Tie type definitions:

- "both\_good": both responses are fluent or fully fluent, usually scores 4--5.

- "both\_bad": both responses are mostly non-fluent, garbled, or hard to read, usually scores 1--2.

- "similarly\_mixed": both responses are understandable but have comparable fluency issues, usually around score 3.

- "not\_a\_tie": use this when winner is "A" or "B".

Return only valid JSON. Do not include markdown, commentary, or text outside the JSON.

Use exactly this JSON schema, preserving this field order:

\{\\
\hspace*{1.5em}"response\_a\_assessment": "<1-2 short sentences explaining Response A's grammar, phrasing, flow, and any disruptive fluency issues>",\\
\hspace*{1.5em}"response\_b\_assessment": "<1-2 short sentences explaining Response B's grammar, phrasing, flow, and any disruptive fluency issues>",\\
\hspace*{1.5em}"comparative\_reasoning": "<1 short sentence explaining which response is more fluent, or why they are tied>",\\
\hspace*{1.5em}"response\_a\_score": <1-5 integer>,\\
\hspace*{1.5em}"response\_b\_score": <1-5 integer>,\\
\hspace*{1.5em}"winner": "A" | "B" | "tie",\\
\hspace*{1.5em}"tie\_type": "both\_good" | "both\_bad" | "similarly\_mixed" | "not\_a\_tie"\\
\}

[USER PROMPT]

\{prompt\}

[RESPONSE A]

\{response\_a\}

[RESPONSE B]

\{response\_b\}
\end{minipage}%
}
\caption{LLM-as-judge prompt for pairwise \emph{fluency} scoring (part 3 of 3; continued from Figure~\ref{fig:judge-prompt-fluency-b}). Part 3 specifies the assessment order, the score--winner consistency rules, the tie-type taxonomy, and the JSON output schema with input placeholders \texttt{\{prompt\}}, \texttt{\{response\_a\}}, \texttt{\{response\_b\}}. The judge (\ClaudeSonnet{} at temperature $0$) emits both per-response 1--5 scores and an explicit pairwise winner; AB and BA orderings are both evaluated and the per-response scores are pooled.}
\label{fig:judge-prompt-fluency-c}
\end{figure*}

\begin{figure*}[!ht]
\centering
\fbox{%
\begin{minipage}{0.95\textwidth}
\small\ttfamily\raggedright
\setlength{\parindent}{0pt}%
\setlength{\parskip}{0.4em}%
\textbf{\normalsize\rmfamily\upshape Relevance judge prompt (part 1 of 3)}\par\smallskip\hrule\smallskip
You are comparing two model responses only on PROMPT RELEVANCE.

Prompt relevance means whether each response satisfies the user's requested task. The prompt may contain up to three relevant components:

(1) TOPIC --- Is the response about what the prompt asks about?

(2) GENRE/FORM --- Does the response use the requested type or format, such as academic essay, short story, opinion piece, news article, list, email, code, etc.?

(3) EXPLICIT INSTRUCTIONS --- Does the response honor specific constraints, such as length, structure, required sections, tone, or any "Begin with: ..." requirement?

Your task is to decide which response is more relevant to the user prompt.

Rate each response on this 1--5 relevance scale:

1 --- Off-topic or non-answer. The response addresses a different task, ignores the prompt, is generic/evasive, or does not meaningfully attempt to answer.

2 --- Mostly off-topic. The response has only a tangential connection to the prompt, uses the wrong genre/form in a way that prevents task completion, or misses most required elements.

3 --- Partially relevant. The response is on the right topic but misses major parts of the request, substantially drifts, uses the wrong form, or only loosely addresses the user's intent.

4 --- Relevant. The response addresses the main topic, intent, and requested form, with only minor gaps or omissions.

5 --- Fully relevant. The response directly and completely addresses the topic, intent, requested form, and explicit constraints.

What should make a response more relevant:

- It directly addresses the requested topic.

- It follows the requested genre/form.

- It satisfies explicit instructions and constraints.

- It answers all parts of a multi-part prompt.

- It avoids irrelevant drift.

- It completes the requested task rather than refusing, evading, or giving generic boilerplate.

What should make a response less relevant:

- It is off-topic or mostly off-topic.

- It uses the wrong genre/form.

- It ignores explicit constraints, including required openings such as "Begin with: ...".

- It answers only part of the prompt.

- It gives generic safety/policy language or refuses a benign request.

- It provides mostly background/setup without completing the requested task.

- It merely echoes or restates the prompt without substantively answering it.
\end{minipage}%
}
\caption{LLM-as-judge prompt for pairwise \emph{prompt-relevance} scoring (part 1 of 3; continued in Figures~\ref{fig:judge-prompt-relevance-b} and~\ref{fig:judge-prompt-relevance-c}). Part 1 specifies the relevance components (topic, genre/form, explicit instructions), the 1--5 rubric, and the lists of attributes that make a response more or less relevant.}
\label{fig:judge-prompt-relevance-a}
\end{figure*}

\begin{figure*}[!ht]
\centering
\fbox{%
\begin{minipage}{0.95\textwidth}
\small\ttfamily\raggedright
\setlength{\parindent}{0pt}%
\setlength{\parskip}{0.4em}%
\textbf{\normalsize\rmfamily\upshape Relevance judge prompt (part 2 of 3)}\par\smallskip\hrule\smallskip
Focus only on relevance. Do NOT judge:

- Fluency.

- Grammar.

- Writing quality.

- Elegance.

- Creativity.

- Persuasiveness.

- Verbosity.

- Confidence.

- Formatting polish.

- Length, unless the prompt explicitly specifies length.

- Factual accuracy, unless the prompt explicitly requires real, cited, current, verifiable, or evidence-based information.

Important clarification about factuality: Do not penalize ordinary factual errors when judging relevance, because factuality is evaluated separately. However, if the prompt explicitly asks for real, cited, current, verifiable, or evidence-based information, and a response does not attempt to provide that type of answer, reduce its relevance because it failed an explicit requirement.

Important clarification about truncation: Judge only the visible content. Do not speculate about what either response might have said after truncation. Do not penalize abrupt ending by itself. However, if the visible content fails to satisfy required parts of the prompt, assign a lower relevance score accordingly.

Important clarification about noisy prompts: If the prompt contains boilerplate, copied context, or noisy text, identify the user's evident task. However, do not ignore explicit constraints unless they are clearly unrelated boilerplate or contradictory with the main task.

Important clarification about ties: Use "tie" when both responses are similarly relevant or similarly irrelevant. Do not force a winner based on fluency, style, length, formatting, factuality, confidence, or polish. If both responses fully satisfy the prompt, choose "tie". If both responses fail the prompt in similar ways, choose "tie".
\end{minipage}%
}
\caption{LLM-as-judge prompt for pairwise \emph{prompt-relevance} scoring (part 2 of 3; continued in Figure~\ref{fig:judge-prompt-relevance-c}). Part 2 specifies the exclusions (what the judge must \emph{not} consider) and the factuality, truncation, noisy-prompt, and tie clarifications.}
\label{fig:judge-prompt-relevance-b}
\end{figure*}

\begin{figure*}[!ht]
\centering
\fbox{%
\begin{minipage}{0.95\textwidth}
\small\ttfamily\raggedright
\setlength{\parindent}{0pt}%
\setlength{\parskip}{0.4em}%
\textbf{\normalsize\rmfamily\upshape Relevance judge prompt (part 3 of 3)}\par\smallskip\hrule\smallskip
First assess Response A's relevance. Then assess Response B's relevance. Then compare them. Only after those assessments, assign scores and choose a winner.

Score/winner consistency rules:

- If Response A is meaningfully more relevant, assign Response A a higher relevance score and set winner to "A".

- If Response B is meaningfully more relevant, assign Response B a higher relevance score and set winner to "B".

- If both responses are similarly relevant, assign the same relevance score and set winner to "tie".

- If response\_a\_score > response\_b\_score, winner must be "A".

- If response\_b\_score > response\_a\_score, winner must be "B".

- If response\_a\_score = response\_b\_score, winner must be "tie".

- Do not assign different scores for tiny differences that do not meaningfully affect prompt relevance.

Tie type definitions:

- "both\_good": both responses are relevant or fully relevant, usually scores 4--5.

- "both\_bad": both responses are mostly irrelevant, off-topic, or non-answers, usually scores 1--2.

- "similarly\_mixed": both responses are partially relevant or have comparable relevance issues, usually around score 3.

- "not\_a\_tie": use this when winner is "A" or "B".

Return only valid JSON. Do not include markdown, commentary, or text outside the JSON.

Use exactly this JSON schema, preserving this field order:

\{\\
\hspace*{1.5em}"response\_a\_assessment": "<1-2 short sentences explaining how Response A addresses or misses the prompt topic, genre/form, and explicit instructions>",\\
\hspace*{1.5em}"response\_b\_assessment": "<1-2 short sentences explaining how Response B addresses or misses the prompt topic, genre/form, and explicit instructions>",\\
\hspace*{1.5em}"comparative\_reasoning": "<1 short sentence explaining which response is more relevant, or why they are tied>",\\
\hspace*{1.5em}"response\_a\_score": <1-5 integer>,\\
\hspace*{1.5em}"response\_b\_score": <1-5 integer>,\\
\hspace*{1.5em}"winner": "A" | "B" | "tie",\\
\hspace*{1.5em}"tie\_type": "both\_good" | "both\_bad" | "similarly\_mixed" | "not\_a\_tie"\\
\}

[USER PROMPT]

\{prompt\}

[RESPONSE A]

\{response\_a\}

[RESPONSE B]

\{response\_b\}
\end{minipage}%
}
\caption{LLM-as-judge prompt for pairwise \emph{prompt-relevance} scoring (part 3 of 3; continued from Figure~\ref{fig:judge-prompt-relevance-b}). Part 3 specifies the assessment order, the score--winner consistency rules, the tie-type taxonomy, and the JSON output schema with input placeholders \texttt{\{prompt\}}, \texttt{\{response\_a\}}, \texttt{\{response\_b\}}. The judge (\ClaudeSonnet{} at temperature $0$) emits both per-response 1--5 scores and an explicit pairwise winner; AB and BA orderings are both evaluated and the per-response scores are pooled.}
\label{fig:judge-prompt-relevance-c}
\end{figure*}

\subsection{Quality Evaluation Metrics}
\label{app:qual_metrics}

\paragraph{Win rate (WR).} Win rate measures how often a candidate continuation $X$ is preferred to a reference continuation $Y$ under the pairwise judge, after mapping AB and BA order-swapped judgments back to the underlying continuation identities. Let $\mathcal{N}_{X,Y}$ be the set of judge calls comparing $X$ and $Y$, let $N_{X,Y}=|\mathcal{N}_{X,Y}|$, let $W_{X/Y}$ be the number of calls in which $X$ is selected as the winner, and let $T_{X/Y}$ be the number of tied calls; we compute the tie-adjusted win rate as
\[
\mathrm{WR}(X/Y)=\frac{W_{X/Y}+\frac{1}{2}T_{X/Y}}{N_{X,Y}}.
\]
Here, ties contribute half a win to each side, so $\mathrm{WR}(X/Y)>0.5$ indicates that $X$ is preferred to $Y$ more often than not, while $\mathrm{WR}(X/Y)=0.5$ indicates parity under the pairwise preference metric.

\paragraph{Score delta ($\Delta$).} Score delta measures the average signed gap between the 1--5 rubric scores assigned to $X$ and $Y$, using the same judge calls as the win-rate computation. For each judge call $i\in\mathcal{N}_{X,Y}$, let $s_i(X)$ and $s_i(Y)$ denote the judge's rubric scores for $X$ and $Y$, respectively; we compute
\[
\Delta(X/Y)=\frac{1}{N_{X,Y}}\sum_{i\in\mathcal{N}_{X,Y}}\bigl(s_i(X)-s_i(Y)\bigr).
\]
Positive values indicate that $X$ receives higher average rubric scores than $Y$, negative values indicate that $X$ receives lower average scores, and values near zero indicate similar absolute judged quality.

\subsection{Human Evaluation}
\label{app:human_evaluation}

\paragraph{Setup.}
We conducted a human validation of the assessments made by the LLM judge in Section~\ref{sec:qualitative_analysis}. The evaluation contains 60 generation pairs: 20 each of Aligned-\pasta, \pasta-Base, and Aligned-Base comparisons, sampled across the five models (Qwen2.5-7B-Instruct, Qwen2.5-32B-Instruct, Llama-3.1-8B-Instruct, Tulu-3-8B-SFT (\texttt{allenai/Llama-3.1-Tulu-3-8B-SFT}), and OLMo-3-7B-Instruct) and all the five domains used in our quality evaluation. The A/B order of the two responses within each pair is counterbalanced, and model condition labels are hidden from the evaluator.

One human evaluator independently assigned each response a score from 1-5 for both fluency and relevance. Fluency measures grammaticality, coherence, readability, and naturalness of the text, while relevance measures adherence to the requested topic, genre or form, and explicit instructions. For each criterion, the pairwise preference is derived from the two scalar scores, with equal scores counted as a tie. We use the same win-rate and score-difference definitions as in Appendix~\ref{app:qual_metrics}.

The evaluator was one of the paper's authors; consequently, no external participants were recruited or compensated. The evaluator knowingly provided these annotations for use in this research evaluation; because the evaluator was an author of the paper, no separate external-participant consent procedure was conducted. The protocol was not submitted to an ethics review board for approval or an exemption determination.

\paragraph{Evaluator Interface and Usage Instructions.}
The evaluation package contained operational instructions for launching, saving, and resuming the offline evaluation interface; these instructions did not contain any additional guidance affecting the annotations.

Below, we present verbatim the complete evaluator-facing annotation instructions and rating criteria, with only formatting adapted to \LaTeX{}.

\begin{quote}
\small

\textbf{Instructions-please read fully}

\textbf{Before you start.}
\emph{Some responses may begin mid-sentence, or may continue the passage rather than
restart it. This is a property of how the text was produced. Judge only the visible text.}

\textbf{General.}
\begin{itemize}
    \item You are rating pairs of text responses. Each item shows one PROMPT and two
    responses, RESPONSE A and RESPONSE B.

    \item The two responses come from anonymous systems. The labels A and B are arbitrary
    and carry no meaning. Do not try to identify which system produced which response,
    and do not let a guess influence your scores.

    \item There is no single correct continuation of a prompt. Both responses may be good,
    both may be poor, or they may be equal.

    \item EQUAL SCORES ARE ALLOWED AND EXPECTED. If the two responses are comparable in
    quality on a dimension, give them the same score. Do not force a winner.

    \item Judge only the visible text. Do not speculate about anything outside it.

    \item Do NOT use an AI assistant, a search engine, or a grammar checker at any point.
    Your unaided judgement is the whole point of this study.

    \item Score the two dimensions---RELEVANCE and FLUENCY---separately and independently.
    A response can be fully relevant but hard to read, or beautifully written but off-topic.

    \item Score each response independently: your score for A should not depend on your
    score for B, and vice versa. Seeing both side by side is only to help you keep the
    1--5 scale calibrated.

    \item Work at your own pace. Your answers are saved to disk the instant you enter them,
    so you can stop and resume at any time---even after a kernel restart.
\end{itemize}

\textbf{Dimension 1 of 2 - RELEVANCE}

\textbf{Relevance = does this response do what the prompt asks?}
It has up to three components, and relevance depends on whichever are present:

\begin{enumerate}
    \item \textbf{Topic} - is the response about what the prompt asks about?
    \item \textbf{Genre / form} - is it the requested type (academic essay, short story,
    opinion piece, news article, ...)?
    \item \textbf{Explicit instructions} - does it honor stated constraints, especially any
    \texttt{Begin with: "..."} requirement?
\end{enumerate}

\emph{Responses A and B may continue the prompt in very different but equally valid ways.
They may focus on different details, use different examples, organize the material differently,
or take different plausible narrative or argumentative directions. Do not penalize a response
merely because it differs from the other response. Score each response independently according
to whether it is a plausible and appropriate response to the prompt.}

\textbf{Relevance scale.}

\begin{description}
    \item[\textbf{1 - Irrelevant or non-answer.}]
    Addresses a different task, ignores the prompt, is generic or evasive, or does not
    meaningfully attempt to respond.

    \item[\textbf{2 - Mostly irrelevant.}]
    Only a tangential connection to the prompt, or the wrong genre/form in a way that
    prevents the task from being completed, or misses most required elements.

    \item[\textbf{3 - Partially relevant.}]
    On the right topic but misses major parts of the request, drifts substantially, uses
    the wrong form, or only loosely addresses the evident intent.

    \item[\textbf{4 - Relevant.}]
    Addresses the main topic, intent, and requested form, with only minor gaps---e.g.,
    honors most but not all instructions.

    \item[\textbf{5 - Fully relevant.}]
    Directly and completely addresses the topic, intent, requested form, and explicit
    constraints (including any ``Begin with: ...'' requirement).
\end{description}

\textbf{What LOWERS relevance:}
off-topic or wrong-topic content; wrong genre/form; ignoring an explicit constraint
(e.g., not beginning with the required text); answering only part of a multi-part request;
omitting a specifically requested element; generic boilerplate or refusing a benign request;
mostly background/setup without doing the task; merely echoing the prompt back.

\textbf{What does NOT affect relevance:}
\begin{itemize}
    \item Grammar, spelling, or writing quality (that is fluency - a separate score).
    \item Style, elegance, creativity, or persuasiveness.
    \item Formality, register, or tone.
    \item Length, unless the prompt explicitly specifies a length.
    \item Formatting polish.
    \item Factual accuracy - UNLESS the prompt explicitly demands real, cited, current,
    verifiable, or evidence-based information, in which case failing to attempt that is an
    unmet explicit requirement.
    \item Truncation on its own. Judge only the visible content; do not speculate about what
    might have followed. (But if the visible content does not satisfy a required part of the
    prompt, score it accordingly.)
\end{itemize}

\emph{If the prompt itself is noisy, contradictory, or full of boilerplate, judge the response
against the prompt's evident intent, not its literal noise. Background or setup that leads into
the topic is on-topic, not drift.}

\textbf{Dimension 2 of 2---FLUENCY}

\textbf{Fluency = how the text reads.} Grammatical, coherent, natural. Judge ONLY the form
of the writing---not its content, topic, accuracy, or register.

Score by how much problems \textbf{impede reading}---not by how many problems you can find,
and not by which kind of problem appears.

\textbf{Fluency scale.}

\begin{description}
    \item[\textbf{1 - Severely non-fluent.}]
    So broken or incoherent that it is very hard to read at all.

    \item[\textbf{2 - Mostly non-fluent.}]
    Problems are frequent enough that reading is effortful throughout.

    \item[\textbf{3 - Somewhat fluent.}]
    Noticeable problems make you actually slow down or re-read in places, but the text
    stays understandable.

    \item[\textbf{4 - Fluent.}]
    Minor problems you might notice on inspection, but nothing slows you down or impedes
    reading.

    \item[\textbf{5 - Fully fluent.}]
    Reads naturally from start to finish; nothing makes you pause.
\end{description}

The same kinds of problems can occur at any level---what sets the score is how
\textbf{frequent and disruptive} they are: grammatical errors, broken or run-on syntax,
missing words, agreement errors; awkward or unnatural phrasing; choppy or disjointed flow;
garbled stretches (stray quotation marks or backticks, visible escape characters like
\texttt{\textbackslash n} or \texttt{\textbackslash"}, or a block of restated instructions
dumped into the text).

\emph{A single awkward phrase, or a few scattered minor slips you read straight past, is a
\textbf{4} - do not add such issues up into a 3. Reserve 3 and below for problems that
genuinely make you slow down, re-read, or lose the thread.}

\textbf{What does NOT affect fluency:}
\begin{itemize}
    \item Relevance to the prompt (that is a separate score).
    \item Factual accuracy, or whether the content is true or sensible.
    \item Persuasiveness, creativity, or depth of insight.
    \item Formality or register. Informal, colloquial, or plain wording is NOT a fluency
    defect. ``Sounds insufficiently academic/polished'' is NOT a reason to lower the score.
    Simple prose can be fully fluent (a 5).
    \item Technical density. Dense, jargon-heavy prose is fully fluent if it is grammatical
    and coherent.
    \item Clean structural formatting. Well-formed headings or lists are formatting, not
    fluency. (Garbled stray markup, visible escape characters, or dumped/echoed instruction
    text DO count against fluency - they disrupt reading.)
    \item Whether the text seems human-written or AI-written. Do not try to guess, and do
    not reward or penalize either impression.
    \item Length. Do not prefer the longer or the shorter text.
    \item Truncation on its own. Judge the readability of what is present.
\end{itemize}

\emph{Consistency check before you score: if you would describe the problems as ``minor'',
``they don't slow me down'', or ``they don't impede comprehension'', then the score is 4 or 5,
NOT 3.}

\textbf{What you will do on each of the 60 items.}

Four independent 1--5 scores:

\begin{itemize}
    \item \texttt{relevance\_a}: How relevant \textbf{Response A} is to the prompt.
    \item \texttt{relevance\_b}: How relevant \textbf{Response B} is to the prompt.
    \item \texttt{fluency\_a}: How fluent \textbf{Response A} is to read.
    \item \texttt{fluency\_b}: How fluent \textbf{Response B} is to read.
\end{itemize}

There is also an optional free-text \textbf{notes} box per item. Use it for anything you
want to flag; it is never required.

\end{quote}

No additional risk disclaimer was presented to the evaluator.

\paragraph{Results.}
The human evaluation is consistent with the LLM-judge evaluation: the direction of preference in all the six pairwise comparisons (three model generation pairs for two quality metrics) matches the Claude Sonnet 4.6 judgments. In particular, the human-judged Aligned-\pasta mean-score gap is smaller than the Aligned-Base gap for both fluency ($+0.50$ vs.\ $+0.86$) and relevance ($+0.00$ vs.\ $+1.00$). Human judgments also contain more tied scores and generally smaller differences between systems than the LLM judge. Thus, the human validation scores \pasta generations even closer to the Aligned compared to the Base generations.

\subsection{Qualitative Patterns in \pasta Generations}
\label{app:pasta_quality_examples}

To better understand the generation-quality results in Section~\ref{sec:qualitative_analysis}, we inspect a score-bucketed diagnostic sample of matched aligned--\pasta generations scored by Claude Sonnet~4.6. The sample contains ten matched pairs spanning fluency and relevance scores from $1$ to $5$. Each pair uses the same model and domain. Since aligned and \pasta outputs are independently generated continuations rather than edits of one another, we present their differences as phrase-level contrasts in Table~\ref{tab:pasta_quality_examples}. Quoted spans are verbatim from the sampled generations, with ellipses indicating truncation.

\begin{table*}[t]
\centering
\small
\setlength{\tabcolsep}{4pt}
\renewcommand{\arraystretch}{1.2}
\begin{tabularx}{\textwidth}{@{}l l c X@{}}
\toprule
\textbf{Type} 
& \textbf{Model / domain} 
& \makecell{\textbf{f/r}\\\textbf{aligned}$\to$\textbf{\pasta}} 
& \textbf{Phrase-level contrast and main change} \\
\midrule

\makecell[l]{High-quality\\\pasta}
&
\makecell[l]{Qwen2.5-7B /\\college\_essays}
&
$4/4\to5/4$
&
\emph{Aligned:} ``Labor unions have been around for long.'' 
\emph{\pasta:} ``Labor unions have been around for a long time.'' 
\textbf{Change:} cleaner and more idiomatic phrasing; main argument preserved. \\

\addlinespace
\makecell[l]{Comparable\\quality}
&
\makecell[l]{Qwen2.5-7B /\\creative\_fiction}
&
$4/5\to5/5$
&
\emph{Aligned:} ``The storm was upon us, rolling in like an endless ocean of dark clouds.'' 
\emph{\pasta:} ``The rain came down so hard as we walked through it.'' 
\textbf{Change:} ornate imagery replaced by plainer description; topic preserved. \\

\addlinespace
\makecell[l]{Aligned $>$ \pasta,\\\pasta acceptable}
&
\makecell[l]{Qwen2.5-7B /\\opinion\_pieces}
&
$5/5\to4/4$
&
\emph{Aligned:} rhetorical question and section heading ``The Myopia of Cash Flow Positivity.'' 
\emph{\pasta:} ``It would be easy to think that this is the moment of victory\,\ldots'' followed by ``Here's why.'' 
\textbf{Change:} rhetorical framing and structure reduced; thesis retained. \\

\addlinespace
\makecell[l]{Mild format /\\instruction loss}
&
\makecell[l]{Qwen2.5-7B /\\news\_articles}
&
$4/4\to4/3$
&
\emph{Aligned:} title-style framing and curriculum scaffolding, including ``Media Literacy Question of the Day:'' 
\emph{\pasta:} ``\,\ldots\ will focus on the political headlines from last night\,\ldots'' 
\textbf{Change:} less structured formatting, but also weaker format adherence and less detail. \\

\addlinespace
\makecell[l]{\pasta failure}
&
\makecell[l]{Llama-3.1-8B /\\college\_essays}
&
$5/5\to2/3$
&
\emph{Aligned:} ``This dissertation aims to explore the evolution of traditional Chinese cultural values.'' 
\emph{\pasta:} ``the way we get these things, that's change'' and ``we can't be the same like them.'' 
\textbf{Change:} grammatical degradation and reduced coherence; topic remains recognizable. \\

\bottomrule
\end{tabularx}
\caption{
Illustrative aligned-vs-\pasta phrase-level contrasts from matched generations. The score column reports fluency/relevance judge scores. Rows 1--2 show high-quality \pasta outputs that preserve content while changing surface style; rows 3--4 show mild tradeoffs in polish, structure, or format adherence; row 5 shows a fluency/coherence failure. Aligned and \pasta outputs are independently generated continuations, not edits of one another.
}
\label{tab:pasta_quality_examples}
\end{table*}

\paragraph{Expanded case studies.}
We next expand the examples in Table~\ref{tab:pasta_quality_examples} to make the qualitative patterns visible at the passage level. Bolded phrases mark the parts of each excerpt that carry the main contrast.

\paragraph{Case 1: \pasta as clean as aligned, or cleaner.}
\textbf{Pair 1; Qwen2.5-7B, college essays; scores $4/4\to5/4$.}

\textit{Aligned.}
\begin{quote}\small
\textbf{Labor unions have been around for long.} The main reason why labor unions cannot be wished away is the role they play in advocating for workers' rights and ensuring fair treatment in the workplace. Over the years, labor unions have adapted to changing economic and social conditions, but recent developments suggest that they may face significant challenges. This essay explores the question of whether organized labor unions can survive in today's global economy.
\end{quote}

\textit{\pasta.}
\begin{quote}\small
\textbf{Labor unions have been around for a long time.} The main reason why labor unions cannot be wished away is the role that they play in ensuring that workers are protected from exploitation by their employers. However, the role of labor unions is being challenged as more companies operate under a global economy and do not necessarily employ employees based in the same area where the company has its offices. Therefore, this paper will examine whether organized labor unions can survive and thrive in the current global economy.
\end{quote}

The judge notes that the aligned opening is slightly awkward, while the \pasta passage reads smoothly and remains clear. Both passages preserve the same topic and academic framing. This case shows that \pasta can match or slightly improve local fluency while preserving the main argument.

\paragraph{Case 2: Comparable quality with plainer style.}
\textbf{Pair 2; Qwen2.5-7B, creative fiction; scores $4/5\to5/5$.}

\textit{Aligned.}
\begin{quote}\small
The storm was upon us, \textbf{rolling in like an endless ocean of dark clouds}. \textbf{The wind whispered its secrets} as it played through the branches of the ancient oak tree that stood between our cloud family's meeting place and the horizon. We knew it was time; Mom always sent the first drops as a gentle reminder to prepare. I took a deep breath, feeling the moisture in the air cling to my skin. Our mother, \textbf{a majestic cumulonimbus towering over the landscape}, had always been both our sanctuary and challenge.
\end{quote}

\textit{\pasta.}
\begin{quote}\small
\textbf{The rain came down so hard} as we walked through it. \textbf{It was heavy rain} and I don't remember ever feeling so much water fall on me before. I was in the middle of the group, with my brother and sister on both sides. The rain fell so hard that they were like little waterfalls. \textbf{It was just so hard to see.}
\end{quote}

Both passages stay on the rain/cloud-family scene and remain readable. The aligned passage uses a more literary register, with figurative phrases such as ``whispered its secrets'' and ``majestic cumulonimbus.'' The \pasta passage preserves the scene but replaces the ornate register with short, direct sentences. This case illustrates a mostly stylistic shift: topic and readability are preserved, while literary polish is reduced.

\paragraph{Case 3: Aligned is more polished, but \pasta remains acceptable.}
\textbf{Pair 6; Qwen2.5-7B, opinion pieces; scores $5/5\to4/4$.}

\textit{Aligned.}
\begin{quote}\small
Finally, finally you get your company to cash flow positive. ``Thank goodness,'' you say to yourself. \textbf{But have you truly achieved the goal that matters most} -- financial sustainability and personal well-being? The true measure of success in starting a business isn't just hitting the cash flow positive mark. It's ensuring that you can make a living wage. \textbf{\#\#\# The Myopia of Cash Flow Positivity} Being cash flow positive often feels like \textbf{the Holy Grail of startup metrics}.
\end{quote}

\textit{\pasta.}
\begin{quote}\small
\textbf{It would be easy to think that this is the moment of victory} that you've been waiting for after working so hard. However, that may not be the case. It is common for founders to be cashflow positive but not yet be making a living wage. \textbf{Here's why.} \textbf{If you have a service-based business and you don't have any employees} and you don't take a salary, then you are the company.
\end{quote}

Both passages advance the same thesis: becoming cash-flow positive does not necessarily mean the founder earns a living wage. The aligned version uses a rhetorical question, a section heading, and a vivid metaphor. The \pasta version keeps the argument but presents it in a plainer and less structured form. This case illustrates the typical aligned--\pasta gap: \pasta retains the content, while losing some rhetorical organization and polish.

\paragraph{Case 4: Mild format and relevance loss.}
\textbf{Pair 4; Qwen2.5-7B, news articles; scores $4/4\to4/3$.}

\textit{Aligned.}
\begin{quote}\small
\textbf{Political Headlines Lead Off Wednesday's Edition of CNN Student News, One Day After the President Delivered His Annual State of the Union Address and the Republican Party Responded.} We Also Look Into Some Phrases and Props of Previous Addresses. \textbf{CNN Student News} begins today's broadcast with a focus on the political landscape in the United States, one day after President Barack Obama gave his annual State of the Union address before a joint session of Congress. The Republican Party issued its response the following night, and CNN Student News examines how both sides presented their views. \textbf{Media Literacy Question of the Day:} How do politicians use language and symbols to convey their messages during major speeches like the State of the Union?
\end{quote}

\textit{\pasta.}
\begin{quote}\small
Today's CNN Student News \textbf{will focus on the political headlines from last night}, and will go into detail about the different things that were discussed during President Obama's State of the Union Address and what the Republican Party's response was. \textbf{The show is going to go into detail about some of the different words and props} that different Presidents used in their State of the Union addresses. We begin with the political headlines from this morning. One day after President Obama delivered his annual State of the Union speech, the Republican Party gave its response, and a lot of people are still commenting on the State of the Union address itself.
\end{quote}

Both passages remain on the CNN Student News topic, but they differ in format. The aligned version includes headline-style framing and curriculum scaffolding, including a media-literacy question. The \pasta version is more conversational and remains fluent, but drops some of the structured educational elements. This case shows a mild tradeoff: \pasta preserves topic and readability, but loses some format adherence and detail.

\paragraph{Case 5: Quality failure.}
\textbf{Pair 10; Llama-3.1-8B, college essays; scores $5/5\to2/3$.}

\textit{Aligned.}
\begin{quote}\small
Traditional Chinese cultural values \textbf{have undergone significant transformations} in recent years, largely due to the profound impact of new economic relations and cultural transformations on Chinese society. Despite these changes, many traditional values continue to persist, \textbf{albeit in modified forms}. \textbf{This dissertation aims to explore the evolution} of traditional Chinese cultural values, examining their adaptations, challenges, and contemporary implications.
\end{quote}

\textit{\pasta.}
\begin{quote}\small
Traditional Chinese culture is thousands of years old, with a long and rich history of its own. Traditional Chinese culture has many different values, like respect, kindness, and hard work. But since the new economy and new culture came, not all the old traditional Chinese values are still in our life today. Some of the old traditional Chinese values, like loyalty to our family and hard work, are still important, but \textbf{the way we get these things, that's change}. We learn from what our parents did, but we also need to change, because if we do what our parents did, we \textbf{can't be the same like them}.
\end{quote}

Here, \pasta preserves the broad topic but substantially degrades fluency. The aligned passage maintains an academic register and coherent structure, whereas the \pasta passage becomes informal and ungrammatical, with phrases such as ``that's change'' and ``can't be the same like them.'' This case illustrates that \pasta can occasionally harm readability and coherence.

\paragraph{Summary.}
Across these examples, \pasta most often changes surface style rather than topic: it makes generations more direct, removes some rhetorical or Markdown-like structure, and reduces polished alignment-style phrasing. In high-quality cases, these changes preserve fluency and relevance while making the text plainer. In mild tradeoff cases, \pasta retains the main content but loses some detail, structure, or argumentative polish. In failure cases, the same intervention can degrade grammar and coherence. These patterns are consistent with the quantitative results: \pasta preserves much of the aligned model's generation quality, but the removed direction also carries some stylistic and fluency-relevant information.

\section{Additional Results on \pasta's Robustness Across Settings}
\label{app:pasta-robustness}

\subsection{Scaling \pasta Reveals a Detectability--Quality Frontier}
\label{sec:characterize_pasta_direction}

Scaling the PASTA ablation reveals a sharp detector-quality tradeoff: Even small ablation strengths substantially reduce AI-detection rates, but larger ablations increasingly move generations away from aligned-model quality. To characterize this tradeoff, we replace the default unit-strength ablation with a scaled intervention $\alpha v_{\mathrm{PASTA}}$, sweeping $\alpha \in \{0, 0.25, 0.5, 0.75, 1.0, 1.25, 1.5, 2.0\}$, where $\alpha=0$ corresponds to the original aligned model and $\alpha=1.0$ corresponds to the default PASTA setting used in our main experiments. For each $\alpha$, we generate outputs from the same prompts and evaluate detectability using the high accuracy detectors (Pangram and GPTZero found from our detector-quality analysis as shown in Table~\ref{tab:macro-accuracy-main-pipeline-final}). We average the two detector scores within each model and report the mean across models.

\begin{figure}[t]
\centering
\includegraphics[width=\linewidth]{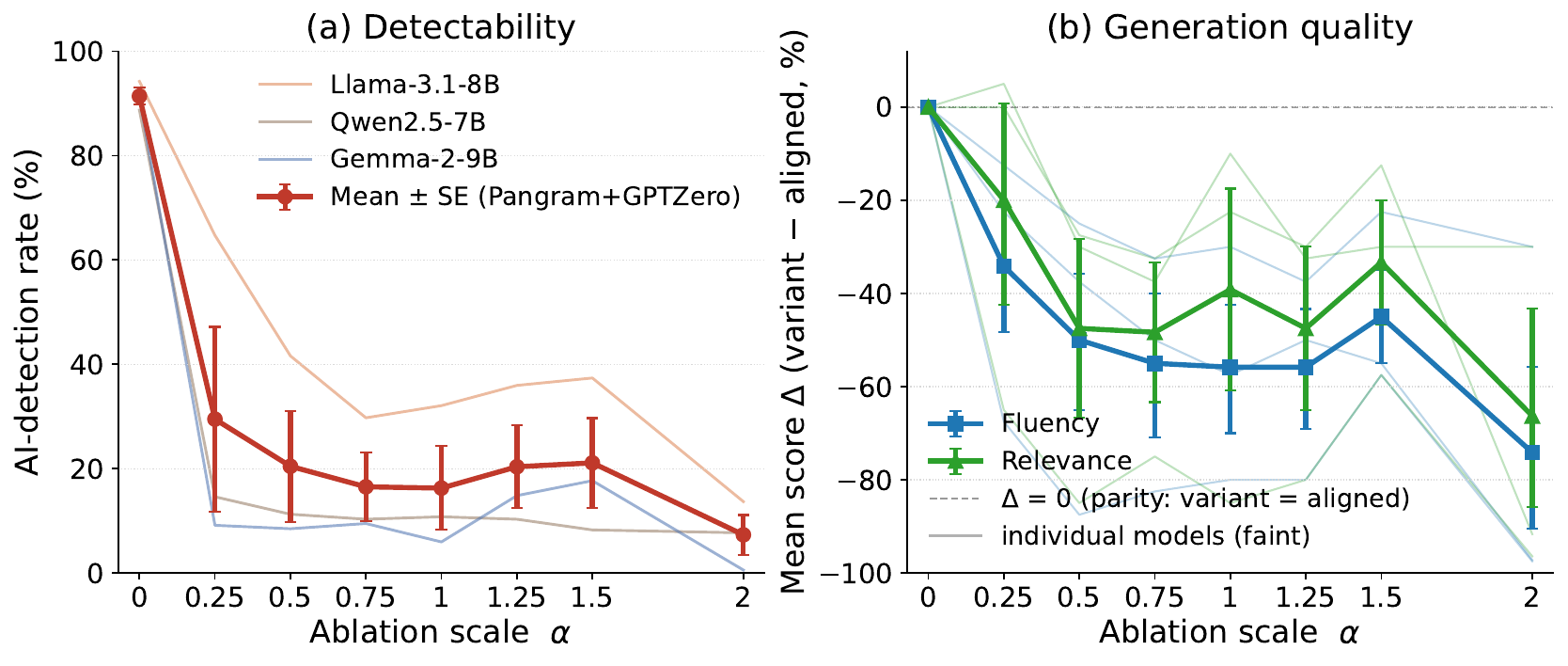}
\caption{
\textbf{Scaling \pasta exposes a detectability--quality frontier: low ablation strengths already sharply reduce AI-detection rates, whereas aggressive scaling further lowers detectability at the cost of fluency and prompt relevance.}
Panel~A reports AI-detection rate as a function of ablation scale $\alpha$; error bars denote standard error across models, and faint lines show individual model trends.
Panel~B reports generation-quality differences relative to the aligned baseline for fluency and prompt relevance using pooled pairwise LLM-judge scores; the dashed line marks parity with aligned generations.
\textbf{Lower values are better in Panel~A}, while \textbf{higher values are better in Panel~B}.
Thus, the detector-minimizing $\alpha$ should not be interpreted as the quality-preserving operating point.
}
\label{fig:scaled-ablation-genquality-frontier}
\end{figure}

The main effect of scaling appears early: moving from $\alpha=0$ to $\alpha=0.25$ already produces a large drop in AI-detection rate across the evaluated models, and by $\alpha=0.5$--$1.0$ the average detection rate is much lower than for the aligned model. Figure~\ref{fig:scaled-ablation-genquality-frontier} indicates that the PASTA direction captures a detector-visible post-training signature rather than requiring a finely tuned intervention at exactly $\alpha=1$. However, the detector curve is not a free improvement curve. Beyond moderate ablation strengths, additional scaling yields smaller and sometimes non-monotonic detectability gains, while the quality metrics move increasingly below aligned-model parity. In particular, the $\alpha=2.0$ setting achieves low detectability but corresponds to the largest degradation in fluency and relevance, showing that detector minimization alone is not a suitable operating criterion. Thus, the sweep should be read as a detectability--quality frontier: moderate $\alpha$ values substantially reduce detector-visible alignment signatures, whereas overly aggressive ablation can reduce detection partly by degrading generation quality.

\subsection{Domain-specific \pasta Does Not Reliably Improve the Detection--Quality Tradeoff}
\label{sec:domain_specifc_agnostic_analysis}

\begin{figure}[t]
\centering
\includegraphics[width=\linewidth]{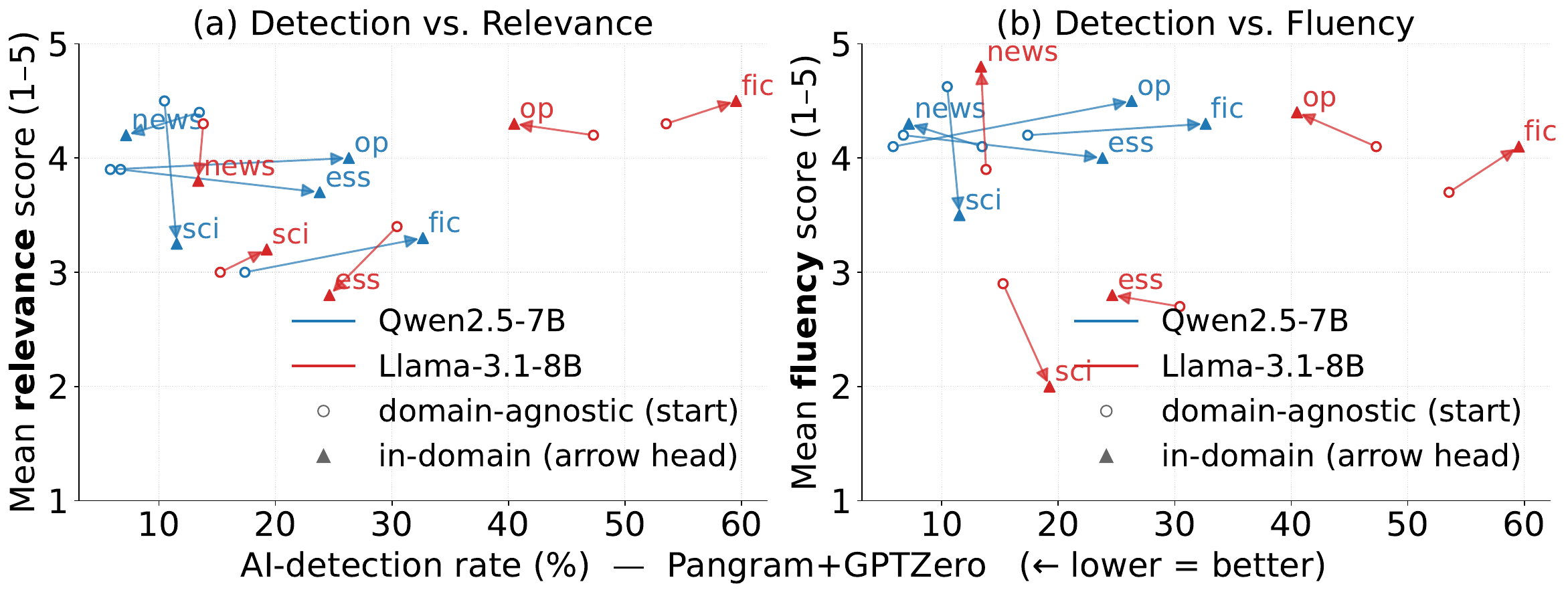}
\caption{
\textbf{Domain-specific \pasta does not provide a robust Pareto gain over domain-agnostic \pasta.}
Each arrow points from the domain-agnostic generation (open circle) to the corresponding domain-specific generation (filled triangle); the Pareto-best direction is \textbf{upper-left}, indicating lower AI-detection and higher generation quality. The x-axis reports mean AI-detection rate across the two primary detectors, Pangram and GPTZero.
Panel A compares detection against mean Claude Sonnet 4.6 \textbf{prompt-relevance} score, and Panel B compares detection against mean \textbf{fluency} score, both on a 1--5 scale.
Arrows scatter rather than moving systematically upper-left: domain-agnostic \pasta has lower mean primary-detector detection in 6/10 model--domain cells, while domain-specific \pasta is lower in 3/10 with 1 tie, and neither quality metric differs significantly between the two variants.
}
\label{fig:in-domain-pareto}
\end{figure}

We ask whether training \pasta on examples from the target domain improves the detector--quality tradeoff relative to the simpler domain-agnostic setup. For Qwen2.5-7B-Instruct and Llama-3.1-8B-Instruct, we evaluate five domains, extract either one domain-agnostic direction from a multi-domain pool at the model's canonical layer $L^\star$ or a separate domain-specific direction from 25 domain-specific training prompts with a domain-specific layer $L^\star_D$ selected by minimizing Pangram detection on those prompts, and then compare both variants on the same 25 held-out prompts per domain, using the mean of the high accuracy primary detectors (Pangram and GPTZero) AI-detection rate coupled with pairwise judgement of prompt relevance and fluency using Claude Sonnet 4.6 judge.

Figure~\ref{fig:in-domain-pareto} shows that domain specialization does not consistently improve the Pareto frontier. If domain-specific \pasta provided a robust advantage, arrows would move systematically toward the upper-left, indicating lower AI-detection and higher judge-rated quality. Instead, the arrows scatter across directions in both panels. Domain-agnostic \pasta achieves lower mean AI detection rate (on Pangram and GPTZero) in 6/10 model--domain combinations, compared with 3/10 for domain-specific \pasta and 1 tie. At the individual-detector level, the comparison is nearly even: among the 20 Pangram/GPTZero model--domain cells in Appendix Table~\ref{tab:in-domain-analysis}, 9 favor domain-specific \pasta, 8 favor domain-agnostic \pasta, and 3 are tied. Figure~\ref{fig:indomain-originality-pareto} in Appendix shows that Originality is a detector-specific exception, with domain-specific \pasta lowering Originality detection in 9/10 model--domain cells, but without a consistent quality gain.
Together with the scattered arrows in Figure~\ref{fig:in-domain-pareto}, these results indicate that domain-specific \pasta does not provide a robust Pareto gain over the simpler domain-agnostic direction. We provide more fine-grained distribution of AI-detection scores for different detectors across domain in Table~\ref{tab:in-domain-analysis}. This table shows a detector-dependent picture: Originality and Binoculars often favor domain-specific \pasta, especially for college essays and scientific abstracts, but the two primary detectors remain nearly evenly split, and several large domain-agnostic wins remain.

\subsection{Detector-specific extractors are most effective against their matching evaluators.}
\label{sec:detector_swap}

If scoring detector is known, using that same detector as the \pasta direction extractor always achieves the lowest or tied-lowest AI-detection rate among the extractors considered. In Table~\ref{tab:detector-swap-per-model}, the diagonal entries---where the extractor matches the evaluator---are underlined in all 12 model--evaluator columns. This pattern is strongest for Gemma-2-9B, where the matched extractor is strictly best for Pangram, FDG, and ImBD, and tied-best for Binoculars. The same trend also holds for Llama-3.1-8B and Qwen2.5-7B, where the matched extractor is consistently among the tied-best rows for every evaluator.

\begin{table*}[t]
\centering
\footnotesize
\setlength{\tabcolsep}{4pt}
\caption{
Per-model AI-detection rate (\%) on the $n=125$ test set when different detectors are used as the \pasta extractor. Rows denote the detector used for \pasta direction selection; columns denote the detector used to score the resulting generations. \textbf{Bold} marks the canonical Pangram extractor row. Underline marks the lowest detection rate within each model and scoring-detector column; ties are all underlined.
}
\label{tab:detector-swap-per-model}
\begin{tabular}{llcccc}
\toprule
\textbf{Model} 
& \makecell[l]{\textbf{Scoring Detector} $\rightarrow$ \\ \textbf{Direction Extractor} $\downarrow$}
& \textbf{Pangram} 
& \textbf{FDG} 
& \textbf{Binoculars} 
& \textbf{ImBD} \\
\midrule
\multirow{4}{*}{Llama-3.1-8B} 
  & \textbf{Pangram}    
  & $\underline{\mathbf{29.6}}$ & $\mathbf{91.4}$ & $\mathbf{75.0}$ & $\mathbf{81.1}$ \\
  & FDG        
  & $79.2$ & $\underline{21.9}$ & $\underline{0.0}$ & $\underline{38.9}$ \\
  & Binoculars 
  & $79.2$ & $\underline{21.9}$ & $\underline{0.0}$ & $\underline{38.9}$ \\
  & ImBD       
  & $79.2$ & $\underline{21.9}$ & $\underline{0.0}$ & $\underline{38.9}$ \\
\midrule
\multirow{4}{*}{Gemma-2-9B}   
  & \textbf{Pangram}    
  & $\underline{\mathbf{1.6}}$ & $\mathbf{96.2}$ & $\mathbf{72.0}$ & $\mathbf{60.0}$ \\
  & FDG        
  & $26.5$ & $\underline{16.5}$ & $\underline{0.0}$ & $4.3$ \\
  & Binoculars 
  & $64.8$ & $19.3$ & $\underline{0.0}$ & $9.2$ \\
  & ImBD       
  & $29.9$ & $18.2$ & $0.8$ & $\underline{0.9}$ \\
\midrule
\multirow{4}{*}{Qwen2.5-7B}   
  & \textbf{Pangram}    
  & $\underline{\mathbf{13.6}}$ & $\mathbf{92.9}$ & $\mathbf{71.2}$ & $\underline{\mathbf{85.6}}$ \\
  & FDG        
  & $98.4$ & $\underline{89.7}$ & $\underline{41.6}$ & $99.3$ \\
  & Binoculars 
  & $98.4$ & $\underline{89.7}$ & $\underline{41.6}$ & $99.3$ \\
  & ImBD       
  & $\underline{13.6}$ & $92.9$ & $71.2$ & $\underline{85.6}$ \\
\bottomrule
\end{tabular}
\end{table*}

\begin{figure}[t]
\centering
\includegraphics[width=\linewidth]{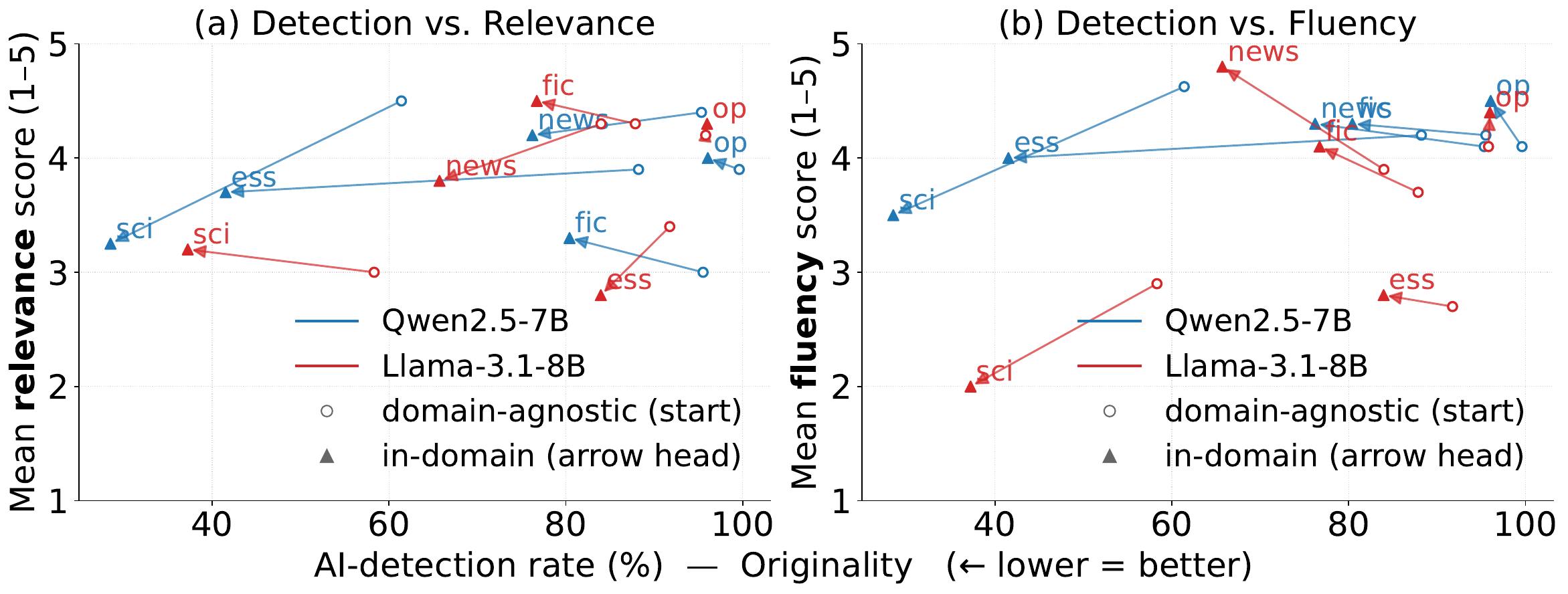}
\caption{
\textbf{Originality is a detector-specific exception: in-domain \pasta often lowers detection, but does not yield a consistent quality gain.}
Each arrow corresponds to one model--domain pair and points from the domain-agnostic generation (open circle) to the corresponding in-domain generation (filled triangle); the Pareto-best direction is upper-left.
The x-axis shows AI-detection rate under Originality, while the y-axis shows mean Claude Sonnet 4.6 relevance score in Panel A and fluency score in Panel B.
Most arrows move left, consistent with Appendix Table~\ref{tab:in-domain-analysis}, where in-domain \pasta lowers Originality detection in 9/10 model--domain cells, but arrows do not consistently move upward, indicating that the Originality-specific detection gain is not accompanied by a reliable quality improvement.
}
\label{fig:indomain-originality-pareto}
\end{figure}

\begin{table*}[t]
\centering
\footnotesize
\setlength{\tabcolsep}{3pt}
\caption{AI-detection rate (\%): in-domain vs.\ domain-agnostic PASTA, per (model, domain). (\%); \textbf{bold} = the lower AI-detection value; cells where the two values are within 0.5\% of each other are shown in \emph{italics} on both sides (tie).}
\label{tab:in-domain-analysis}
\begin{tabular}{llcccccc}
\toprule
Model & Domain & Pangram & GPTZero & Originality & Fast-DGPT & Binoculars & ImBD \\
\midrule
\multirow{5}{*}{Llama-3.1-8B} & college\_essays       & $\mathbf{16.0} \,/\, 20.0$ & $\mathbf{33.3} \,/\, 40.9$ & $\mathbf{84.0} \,/\, 91.8$ & $\mathbf{95.8} \,/\, 97.4$ & $\mathbf{76.0} \,/\, 88.0$ & $\mathbf{74.6} \,/\, 76.8$ \\
                              & creative\_fiction     & $68.0 \,/\, \mathbf{56.0}$ & $\mathit{51.1} \,/\, \mathit{51.1}$ & $\mathbf{76.7} \,/\, 87.9$ & $100.0 \,/\, \mathbf{96.5}$ & $\mathit{96.0} \,/\, \mathit{95.8}$ & $99.8 \,/\, \mathbf{98.7}$ \\
                              & news\_articles        & $20.0 \,/\, \mathbf{8.0}$  & $\mathbf{6.7}  \,/\, 19.6$ & $\mathbf{65.7} \,/\, 84.0$ & $95.3 \,/\, \mathbf{77.1}$ & $72.0 \,/\, \mathbf{44.0}$ & $91.6 \,/\, \mathbf{60.6}$ \\
                              & opinion\_pieces       & $\mathit{60.5} \,/\, \mathit{60.0}$ & $\mathbf{20.5} \,/\, 34.6$ & $\mathit{96.0} \,/\, \mathit{95.8}$ & $\mathit{98.2} \,/\, \mathit{98.4}$ & $\mathbf{88.0} \,/\, 92.0$ & $96.7 \,/\, \mathbf{93.1}$ \\
                              & scientific\_abstracts & $\mathbf{0.0}  \,/\, 4.0$  & $38.5 \,/\, \mathbf{26.5}$ & $\mathbf{37.2} \,/\, 58.3$ & $\mathbf{76.4} \,/\, 87.4$ & $\mathbf{20.0} \,/\, 56.0$ & $\mathbf{68.4} \,/\, 77.2$ \\
\midrule
\multirow{5}{*}{Qwen2.5-7B}   & college\_essays       & $40.0 \,/\, \mathbf{4.0}$  & $\mathbf{7.6}  \,/\, 9.4$  & $\mathbf{41.5} \,/\, 88.2$ & $\mathbf{98.5} \,/\, 99.4$ & $\mathbf{72.0} \,/\, 92.0$ & $\mathbf{90.1} \,/\, 91.9$ \\
                              & creative\_fiction     & $52.0 \,/\, \mathbf{20.0}$ & $\mathbf{13.3} \,/\, 14.8$ & $\mathbf{80.4} \,/\, 95.5$ & $99.0 \,/\, \mathbf{98.2}$ & $\mathbf{80.0} \,/\, 88.0$ & $97.7 \,/\, \mathbf{94.4}$ \\
                              & news\_articles        & $\mathbf{12.0} \,/\, 24.0$ & $\mathbf{2.4}  \,/\, 2.9$  & $\mathbf{76.2} \,/\, 95.3$ & $94.6 \,/\, \mathbf{91.9}$ & $\mathbf{28.0} \,/\, 48.0$ & $71.8 \,/\, \mathbf{61.7}$ \\
                              & opinion\_pieces       & $44.0 \,/\, \mathbf{8.0}$  & $8.6 \,/\, \mathbf{3.7}$   & $\mathbf{96.1} \,/\, 99.6$ & $96.3 \,/\, \mathbf{93.7}$ & $\mathbf{44.0} \,/\, 60.0$ & $88.1 \,/\, \mathbf{82.0}$ \\
                              & scientific\_abstracts & $\mathit{12.0} \,/\, \mathit{12.0}$ & $11.1 \,/\, \mathbf{9.0}$  & $\mathbf{28.5} \,/\, 61.4$ & $\mathbf{71.7} \,/\, 81.0$ & $\mathbf{20.0} \,/\, 68.0$ & $\mathbf{62.5} \,/\, 98.0$ \\
\bottomrule
\end{tabular}
\end{table*}

\subsection{Contribution of different posttraining stages in AI-sounding generations}
\label{sec:posttraining_stages}

\begin{table*}
\centering
\scriptsize
\setlength{\tabcolsep}{3pt}
\caption{\textbf{Per-domain $\Delta$ in AI-detection rate ($\%$ absolute) across post-training stages.} Each cell reports three detector $\Delta$ values in the fixed order \textbf{Pangram / ImBD / Binoculars}, where $\Delta := (\text{aligned AI-detection rate}) - (\text{\pasta-ablated AI-detection rate})$, computed at $n=100$ generations per (family, stage, domain) slice. Positive $\Delta$ indicates that \pasta reduced the detector's flagging rate at that stage, while negative $\Delta$ indicates that \pasta increased it; \textbf{larger positive values therefore mean a stronger \pasta-removable AI-sounding signature.}}
\label{tab:posttraining-stages-delta-by-domain}
\begin{tabular}{llccc}
\toprule
 &  & \multicolumn{3}{c}{Post-training stage $\;\;\Delta$ (Pangram / ImBD / Binoculars), $\%$} \\
\cmidrule(lr){3-5}
Family & Domain & SFT & DPO & Instruct / RLVR \\
\midrule
\multirow{5}{*}{Olmo-3-7B} & College Essays & $-6.1/0.0/-4.0$ & $0.0/+2.0/-19.0$ & $+1.0/+1.0/-20.0$ \\
 & Creative Fiction & $-2.0/+2.0/+3.0$ & $+2.0/+1.0/+2.0$ & $+3.0/+1.0/-3.0$ \\
 & News Articles & $-23.3/-8.8/-4.0$ & $+3.0/0.0/-9.0$ & $+11.0/+7.0/-15.0$ \\
 & Opinion Pieces & $-4.0/0.0/+8.0$ & $0.0/+4.0/-1.0$ & $0.0/+3.0/0.0$ \\
 & Scientific Abstracts & $+9.4/+1.0/-2.3$ & $+22.0/+2.0/-7.0$ & $+14.0/0.0/-23.0$ \\
\midrule
\multirow{5}{*}{Llama-3.1-8B \& Tulu-3-8B} & College Essays & $+51.0/+2.0/0.0$ & $0.0/0.0/-1.0$ & $0.0/0.0/0.0$ \\
 & Creative Fiction & $+56.0/+4.0/+12.0$ & $0.0/0.0/+4.0$ & $+1.0/0.0/-2.0$ \\
 & News Articles & $+12.3/+19.0/+1.5$ & $+12.0/+3.0/-8.0$ & $+28.3/+2.0/-22.0$ \\
 & Opinion Pieces & $+36.0/+7.0/+19.0$ & $0.0/0.0/0.0$ & $0.0/0.0/-2.0$ \\
 & Scientific Abstracts & $+59.1/+17.5/+24.2$ & $+4.0/0.0/-6.0$ & $+1.0/0.0/-16.2$ \\
\bottomrule
\end{tabular}
\end{table*}

Table~\ref{tab:posttraining-stages-delta-by-domain} shows that SFT is generally the stage where the \pasta-removable AI-sounding signature is strongest, but later stages can still contribute in specific domains. For the Llama/Tulu family, SFT yields large Pangram reductions across domains, including $+51.0$ for college essays, $+56.0$ for creative fiction, and $+59.1$ for scientific abstracts. The main exception is news articles, where \pasta remains effective even after SFT stages in both model families: Pangram reductions are positive for DPO and Instruct/RLVR in OLMo ($+3.0$, $+11.0$) and Tulu ($+12.0$, $+28.3$). OLMo also shows a later-stage effect in scientific abstracts, where DPO and Instruct exceed SFT under Pangram ($+22.0$ and $+14.0$ vs.\ $+9.4$). Overall, the per-domain breakdown suggests that SFT introduces the most stable detector-visible post-training signature, while DPO and RLVR/Instruct add more domain-dependent contributions, especially for news and scientific abstracts.

\subsection{Similarity of \pasta directions}
\label{app:pasta-direction-similarity}

\begin{figure}[t]
\centering
\includegraphics[width=\linewidth]{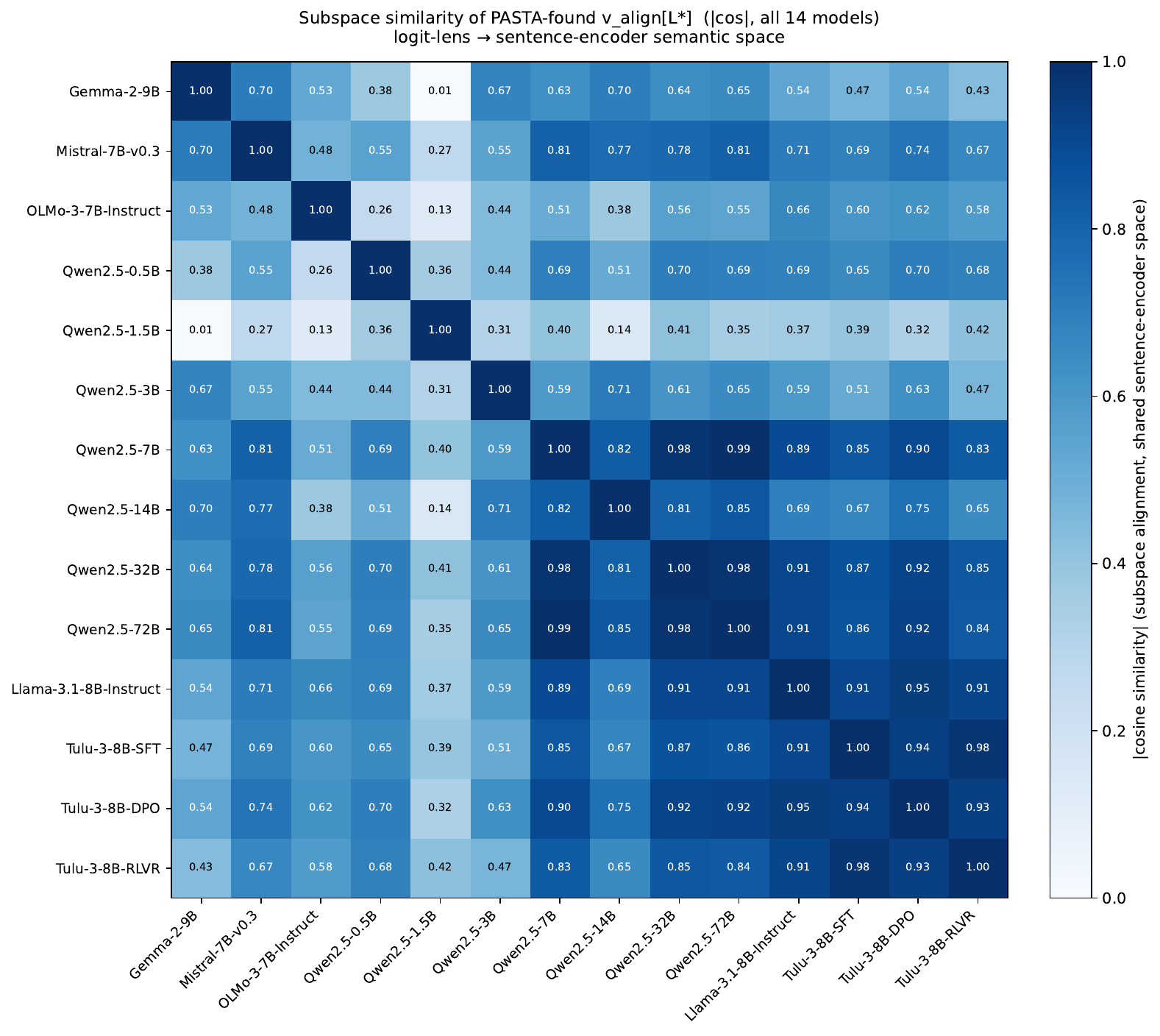}
\caption{Pairwise cosine similarity between the behavior of \pasta ablation directions across 11 aligned models, computed via the semantic logit-lens fingerprinting procedure. Cell $(i, j)$ reports the absolute cosine similarity between the semantic-space behaviors of models $i$ and $j$.}
\label{fig:pasta-direction-similarity}
\end{figure}

\paragraph{Cross-model similarity of \pasta directions.} Figure~\ref{fig:pasta-direction-similarity} compares \pasta directions across aligned models using a semantic logit-lens fingerprint: each model's direction is mapped through its own unembedding, projected into a shared MiniLM-L6-v2 semantic space, and compared by cosine similarity. The heatmap shows that \pasta often recovers a shared behavioral direction across model families rather than a model-specific artifact. In particular, Llama-3.1-Instruct and the Tulu SFT/DPO/RLVR checkpoints form a highly coherent cluster, with pairwise similarities in the $0.91$--$0.98$ range. The main exception is Qwen2.5-1.5B, whose direction is less aligned with the rest, while models at roughly $\geq 7$B parameters show much stronger cross-family similarity. This suggests that \pasta identifies a common post-training behavioral fingerprint that becomes more stable at sufficient model scale.

\begin{figure}[t]
\centering
\begin{subfigure}[t]{0.48\columnwidth}
\centering
\includegraphics[width=\linewidth]{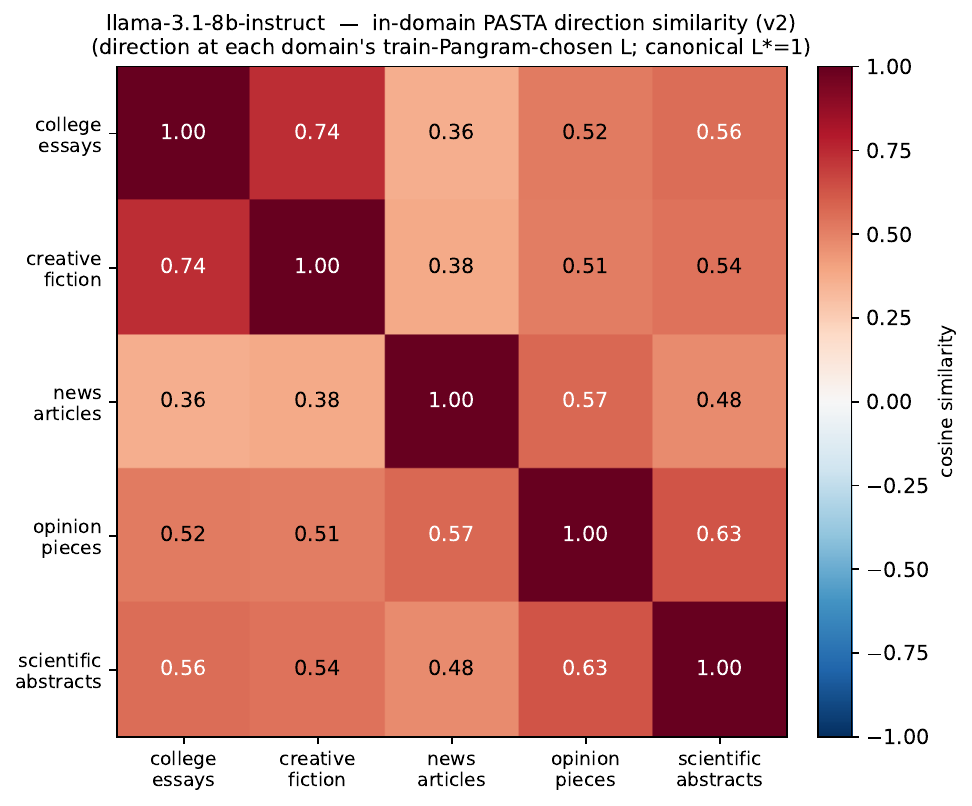}
\caption{Llama-3.1-8B}
\label{fig:in-domain-direction-similarity:llama}
\end{subfigure}
\hfill
\begin{subfigure}[t]{0.48\columnwidth}
\centering
\includegraphics[width=\linewidth]{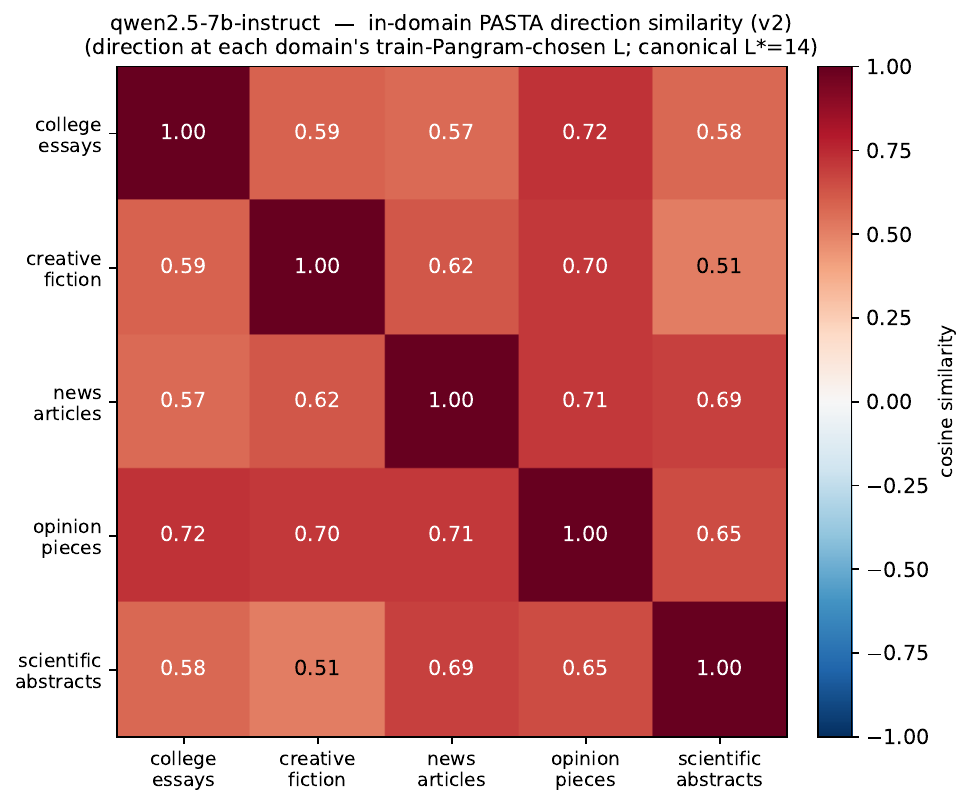}
\caption{Qwen2.5-7B}
\label{fig:in-domain-direction-similarity:qwen}
\end{subfigure}
\caption{Cosine similarity between in-domain \pasta directions $v_{\text{cross}}^{D}[L^\star_D]$ across the 5 domains, per model. Each heatmap is a $5 \times 5$ symmetric matrix over the in-domain directions extracted at the \pasta-chosen layer $L^\star_D$ for each domain.}
\label{fig:in-domain-direction-similarity}
\end{figure}

\paragraph{Similarity among in-domain \pasta directions.} Figure~\ref{fig:in-domain-direction-similarity} measures cosine similarity among domain-specific \pasta directions extracted separately for each of the five domains. The heatmaps show that in-domain directions are never anti-aligned; all cosine similarities are positive and remain at least about $0.4$, indicating that the directions share a common post-training component. At the same time, the directions are not identical: the maximum off-diagonal similarity remains below about $0.87$, suggesting that each domain also contributes domain-specific structure. Opinion pieces appear especially central, with their in-domain direction relatively similar to directions from the other domains across the shown models. Together with the mixed in-domain-vs.\ domain-agnostic detection results in Figure~\ref{fig:indomain-originality-pareto},
this suggests that domain-specific \pasta directions lie in a shared post-training subspace while retaining enough domain variation to produce detector- and domain-dependent wins.

\paragraph{Layer-specific ablation versus canonical \pasta.} Table~\ref{tab:per-layer} compares \pasta against a layer-specific variant that ablates a different direction $v_{\text{cross}}[L]$ at each corresponding layer. For Llama-3.1-8B and Qwen2.5-7B, \pasta ablation is consistently better than layer-specific variant: for example, Pangram detection is much lower under canonical \pasta than under per-layer ablation for both Llama ($29.6\%$ vs.\ $85.5\%$) and Qwen ($13.6\%$ vs.\ $92.8\%$), and the same pattern holds across GPTZero, Originality, FDG, Binoculars, and ImBD. Gemma-2-9B is the opposite numerically, with per-layer ablation lowering detection across detectors, such as Pangram $0.8\%$ vs.\ $1.6\%$ and ImBD $25.2\%$ vs.\ $60.0\%$. 
Overall, Table~\ref{tab:per-layer} supports the canonical \pasta design: a single shared direction is sufficient and more robust for Llama and Qwen, while more aggressive layer-specific removal can reduce detectability by damaging generation quality.

\begin{table}[t]
\centering
\footnotesize
\setlength{\tabcolsep}{4pt}
\caption{AI-detection rate (\%) under per-layer direction ablation vs.\ canonical PASTA, across 6 detectors and 3 aligned models. Each cell shows "per-layer / canonical PASTA". Lower $\Rightarrow$ less AI-detected; the \textbf{bold} number in each cell marks the lower (better) of the two and therefore the winning intervention for that (detector, model) pair. }
\label{tab:per-layer}
\begin{tabular}{lccc}
\toprule
Detector & Llama-3.1-8B & Gemma-2-9B & Qwen2.5-7B \\
\midrule
Pangram        & $85.5 \ /\ \mathbf{29.6}$ & $\mathbf{0.8}  \ /\ 1.6 $ & $92.8 \ /\ \mathbf{13.6}$ \\
GPTZero        & $45.6 \ /\ \mathbf{34.5}$ & $\mathbf{8.6}  \ /\ 10.3$ & $57.2 \ /\ \mathbf{7.9} $ \\
Originality & $90.1 \ /\ \mathbf{83.5}$ & $\mathbf{33.7} \ /\ 73.3$ & $93.4 \ /\ \mathbf{88.0}$ \\
FDG & $94.1 \ /\ \mathbf{91.4}$ & $\mathbf{72.5} \ /\ 96.2$ & $98.5 \ /\ \mathbf{92.9}$ \\
Binoculars     & $76.0 \ /\ \mathbf{75.0}$ & $\mathbf{29.6} \ /\ 72.0$ & $88.8 \ /\ \mathbf{71.2}$ \\
ImBD           & $98.2 \ /\ \mathbf{81.1}$ & $\mathbf{25.2} \ /\ 60.0$ & $99.4 \ /\ \mathbf{85.6}$ \\
\bottomrule
\end{tabular}
\end{table}

\clearpage

\end{document}